\documentclass[11pt]{article}

\usepackage[final]{acl}

\usepackage{times}
\usepackage{latexsym}
\usepackage{bbm}
\usepackage[utf8]{inputenc}
\usepackage{longtable}
\usepackage{array}
\usepackage{colortbl}
\usepackage{xcolor}
\usepackage{diagbox}
\usepackage{float}
\usepackage{enumitem}
\usepackage{soul}
\usepackage{graphicx}
\usepackage{subcaption}
\usepackage{amsfonts}
\usepackage{amsmath}
\usepackage{mathrsfs}
\usepackage{booktabs}
\usepackage{xargs}
\usepackage{csvsimple}
\usepackage{booktabs}
\usepackage{siunitx} 

\usepackage[T1]{fontenc}

\usepackage{microtype}

\usepackage{inconsolata}

\usepackage[colorinlistoftodos,prependcaption,textsize=small,disable]{todonotes}

\title{Give it Space! Explicit Disentangling of Positional and Semantic Representations in Encoders}

\author{
 \textbf{Pierre-Antoine Lequeu\textsuperscript{1}}, \
 \textbf{Camille Barboule\textsuperscript{2}}, \ 
 \textbf{Benjamin Piwowarski\textsuperscript{1}}
\\
 \textsuperscript{1} Sorbonne Université, CNRS, ISIR, Paris, France \\
 \textsuperscript{2} Orange Innovation \\
 \small{
   \textbf{Correspondence:} \href{mailto:lequeu@isir.upmc.fr}{lequeu \textit{(at)} isir.upmc.fr}
 }
}
\begin{document}
\maketitle
\begin{abstract}

Positional encoding (PE) underpins how permutation-invariant Transformers represent sequence order, yet how positional information is processed and stored remains poorly understood. Modern PE methods such as RoPE still struggle on tasks such as long-context understanding or retrieval \cite{chen-etal-2025-hope}. Hence, a better understanding of the internal positional mechanism could help design better PE. Building on evidence that positional and semantic signals occupy nearly orthogonal subspaces in trained Transformers, we modify an encoder Transformer to process three explicitly disentangled streams: semantic, absolute positional (AP) and relative positional (RP), and confine the masked-language-modeling (MLM) objective to the semantic stream. This decoupling enables a clean mechanistic study and yields three take-aways. (1) The isolated AP subspace spontaneously collapses into a low-frequency two-dimensional manifold that captures the structure of the document; (2) Attention heads specialize into structure and semantic-oriented groups, with RP exclusively supporting the latter; (3) Standard positional encodings do not robustly retain macroscopic structure: RoPE and RP only weakly encode it, and entangled AP loses it in the final layers under MLM pressure. The disentangled approach preserves positional encoding, which improves linguistic representations.

\end{abstract}

\section{Introduction}
\todo[inline]{relecture (avec abstract)}


The shift from absolute to relative positional encoding discarded something valuable. Modern Transformers increasingly rely on Relative Positional Encoding (RPE), via additive attention biases \cite{raffel-etal-2020-exploring, press-etal-2022-train, chi-etal-2022-kerple, li-etal-2024-functional} or rotary transformations (RoPE) \cite{su-etal-2023-roformer}, which encodes position only during attention computation. Unlike earlier absolute positional (AP) embeddings \cite{vaswani-etal-2017-attention, devlin-etal-2019-bert}, RPE leaves no persistent positional trace in hidden representations. Moreover, RPE methods often rely on long-term decay heuristics that assume attention diminishes with distance, creating positional biases \cite{wu-etal-2025-emergence} and bottlenecking long-context tasks \cite{chen-etal-2025-hope}. \citet{gu2026deconstructing} further demonstrated that this forced positional dependence is often unnecessary; many heads function perfectly well without positional signals. 
Therefore, the fundamental question of how positional information is used, and how to best represent it, remains open across architectures. Given recent evidence that models learn positional and semantic signals in nearly orthogonal subspaces \cite{song-etal-2024-uncovering}, explicitly separating these signals would provide new insights into the internal mechanisms used in models and help define better position representations in transformers. Additionally, providing separated representations could allow for richer embeddings, since they do not need to mix information. We present in this work the first mechanistic exploration of such a disentangled system to tackle the following research questions:  

\begin{figure}[t]
\centering
\includegraphics[width=1\linewidth]{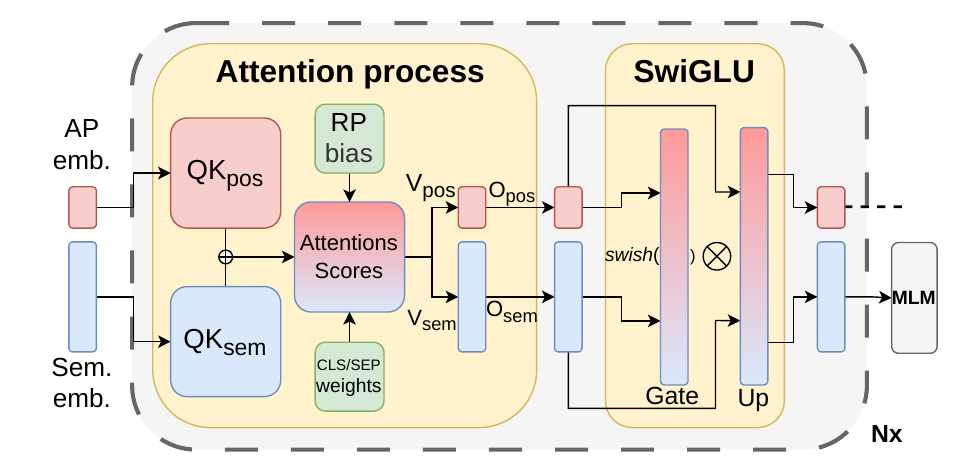}
\caption{The disentangled architecture. Absolute positional (AP) embeddings are displayed in red, and semantic embeddings in blue. Red-blue gradient components contain shared information. Separate RMSNorms are not displayed for readability.}
\label{fig:posbert_architecture}
\end{figure}


\begin{itemize}[itemsep=0pt, parsep=1pt, partopsep=0pt, topsep=0pt]
    \item Does explicitly disentangling positional and semantic streams provide insights into Transformers' internal mechanisms?
    \item What geometric structure and functional taxonomy emerge within the network when positional and semantic information streams are separated?
    \item Does explicitly offloading position to an isolated subspace yield richer representations?
\end{itemize}

We compare three baselines, one using RoPE, one using learned AP embeddings, and one using a learned RP bias, with a disentangled architecture using separated AP embeddings and a learned RP bias. 
After introducing the architectural changes in Section~\ref{sec:posbert_architecture}, we make the following contributions:
\begin{itemize}[itemsep=0pt, parsep=1pt, partopsep=0pt, topsep=0pt]
    \item Through a mechanistic exploration, we reveal that the isolated AP space collapses into a two-dimensional, low-frequency structural manifold (94\% variance in 2 principal components), and that attention heads specialize almost exclusively into structure-oriented or semantic-oriented groups (Section~\ref{sec:analysis}).
    \item In our experimental setting, encoders with RPE methods do not keep structural information in their representations. While entangled AP embeddings can encode this structural information, they lose most of it in the final layer due to interference from the MLM prediction objective. Disentangling the streams solves this positional bottleneck.
    \item We demonstrate that this disentanglement enhances the linguistic fidelity of the token representations: it improves probing results on 42 of the 65 linguistic phenomena of the Flash-Holmes probing benchmark.
\end{itemize}
For reproducibility and further research, we share our implementation and experimental scripts.\footnote{https://github.com/LequeuISIR/DSTG-encoder}

\section{Related Works}
\paragraph{Positional Encoding}
Positional encoding (PE) has evolved from absolute positional embeddings (APE) --- fixed sinusoidal \cite{vaswani-etal-2017-attention} or learned \cite{devlin-etal-2019-bert, liu-etal-2020-learningencode} --- added directly to token representations, to relative positional encoding (RPE) injected into attention logits. Additive RPE methods include T5’s bucketed bias \cite{raffel-etal-2020-exploring}, ALiBi’s fixed decay \cite{press-etal-2022-train}, and refinements such as KERPLE \cite{chi-etal-2022-kerple}, FiRE \cite{li-etal-2024-functional} and Sandwich \cite{chi-etal-2023-dissecting}. Rotary Positional Encoding (RoPE) \cite{su-etal-2023-roformer} applies multiplicative rotations to keys and queries, with extensions like YaRN \cite{peng-etal-2023-yarn} improving long-context extrapolation. All these RPE methods share a key property: positional information is consumed during attention and does not persist in hidden states.

\paragraph{Semantic-Positional Interaction in PE}
Several approaches address the interaction between positional and semantic information. Some implicitly modify RoPE to better handle long inputs: HoPE \cite{chen-etal-2025-hope} removes the low-frequency components of RoPE to mitigate negative biases in long contexts, and PoPE \cite{gopalakrishnan-etal-2026-decoupling} removes the effect of the semantic embeddings on the phases of RoPE's rotations. Other works explicitly condition the positional information on the semantic information:  BiPE \cite{he-etal-2024-two} uses dual encodings for inter-segment and intra-segment positions. CoPE \cite{golovneva-etal-2024-contextual} learns a gating mechanism to assign context-dependent positional encodings. DaPE \cite{zheng-etal-2024-dape, zheng-etal-2025-dape} conditions its positional bias on semantic information through a linear transformation. While these PE methods better account for the positional-semantic interaction, they do not allow direct access to each type of information. 
Our architecture shares TUPE's \cite{ke-etal-2021-rethinking} separation of AP and RP, but replaces its static, shared AP bias with an evolving AP subspace that progressively refines structural representations and mixes with semantic space in the feed-forward network. Our work is also closely related to RePo \cite{li-etal-2026-repo} which extracts positional information from embeddings, and uses head-dependent projections to get position indices for each token. However, RePo does not aim to separate the two information streams, and does not provide a mechanistic analysis of the learned representations.   

\paragraph{Positional Space in Transformers} 
Several studies examine latent space geometry. \citet{wang-chen-2020-position} show that learned APEs in models like BERT \cite{devlin-etal-2019-bert} and RoBERTa \cite{liu-etal-2019-roberta} are high-dimensional because they conflate positional and structural information. \citet{song-etal-2024-uncovering} reveal that positional and semantic bases are nearly orthogonal, where nearby-token attention is driven by position and matching behaviors by content. Positional processing is not uniformly distributed; \citet{gu2026deconstructing} show it often concentrates in a few shallow specialized heads, while \citet{urrutia2025decouplingpositionalsymbolicattention} observe a layerwise shift toward more symbolic heads in deeper layers. Relatedly, \citet{wu-etal-2024-retrieval} identify \emph{retrieval heads} that bypass positional information entirely to perform purely semantic matching. Our disentangled design makes this specialization explicit.



\section{Architecture and Training}
\label{sec:posbert_architecture}

To explore the disentanglement of positional and semantic information, each component of the architecture must be modified. In this section, we introduce the architecturally separated semantic, absolute-positional, and relative-positional information streams. We base our architecture on NeoBERT \cite{breton-etal-2025-neobert} and we describe below the modifications to each component. The full architecture is displayed in Figure~\ref{fig:posbert_architecture}.


\subsection{Attention Blocks}
The attention blocks process the three information streams to produce jointly computed attention scores, which are then applied to both the AP and semantic value streams.

\paragraph{Input Token Representations}
Each token is represented by two embeddings: a $d_{AP}$-dimensional AP embedding and a $d_{sem}$-dimensional semantic embedding. We define the hidden size of the model as $d_{model}=d_{AP} + d_{sem}$. We use the \texttt{bert-base-uncased} tokenizer.

\paragraph{RMSNorm}
The pre-attention and post-attention layer norms, using the RMS norm \cite{Zhang-etal-2019-root}, are separated between the AP and the semantic embeddings to avoid creating an inter-dependence between the two.

\paragraph{Relative Positional Bias} Many recent biases \cite{press-etal-2022-train, chi-etal-2022-kerple, li-etal-2024-functional} rely on kernels parameterized by the signed distance $i-j$, which imposes a long-term decay prior. To avoid baking such a prior into our analysis, we instead adopt the bucketed bias introduced in T5 \cite{raffel-etal-2020-exploring} (detailed in Appendix~\ref{app:t5-positional-bias}), which offers higher precision for nearby tokens but becomes increasingly coarse as the distance between tokens increases, while leaving the decay behaviour to be learned. Each bucket has its own learned parameter $\rho^h_{\text{bucket}(i-j)}$, independent of its distance to the attending token. Additionally, following TUPE \cite{ke-etal-2021-rethinking}, we argue that the position of special tokens $\mathscr{S}=\{\text{[CLS]}, \text{[SEP]}\}$ at the beginning and the end of a sequence is purely arbitrary and should not be used to compute their relative attention weights. Therefore, their RP bias corresponds to head-specific distance-agnostic learned parameters: $\rho^h_{i \leftarrow j}$ when both tokens are special, $\rho^h_{\leftarrow j}$ when attending \emph{from} a special token $j$ to a regular token, and $\rho^h_{i \leftarrow}$ when a regular token attends \emph{to} a special token $i$. The RP bias in head $h$ is:

\begin{equation}
  b^h_{i\leftarrow j} =
    \begin{cases}
      \rho^h_{i \leftarrow j} & \text{if } i \in \mathscr{S} \wedge j \in \mathscr{S}\\
      \rho^h_{\leftarrow j} & \text{if } j \in \mathscr{S}\\
      \rho^h_{i \leftarrow} & \text{if } i \in \mathscr{S}\\
      \rho^h_{\text{bucket}(i-j)} & \text{otherwise}
    \end{cases}
\end{equation}


\paragraph{Attention Heads} For multi-head attention with $n_{heads}$ heads, queries and keys are computed independently for each stream within each head $h$: $Q^h_{sem} = W_{q}^{h,sem} x_{sem}$, $K^h_{sem} = W_{k}^{h,sem} x_{sem}$, $Q^h_{AP} = W_{q}^{h,AP} x_{AP}$, $K^h_{AP} = W_{k}^{h,AP} x_{AP}$, where $W_{q}^{h,AP}, W_{k}^{h,AP} \in \mathbb{R}^{d_{head} \times d_{AP}}$ and $W_{q}^{h,sem}, W_{k}^{h,sem} \in \mathbb{R}^{(d_{head}) \times d_{sem}}$, with $d_{head} = d_{model} / n_{heads}$. A key design choice: we project both streams to the same per-head size $d_{head}$ despite the different input sizes ($d_{AP}$ and $d_{sem}$), ensuring that the dot-product variances are matched and preventing the higher-dimensional space from dominating the attention computation. This requires $d_{AP}$ and $d_{sem}$ to be multiples of $n_{heads}$; in our configuration $d_{AP}/n_{heads} = 8$, which is small but suffices since the AP stream carries low-bandwidth structural information. The attention weight matrices are $w^{h,sem}_{i \leftarrow j} = \frac{(Q^h_{sem})_i \cdot (K^h_{sem})_j}{\sqrt{d_{head}}}$ and $w^{h,AP}_{i \leftarrow j} = \frac{(Q^h_{AP})_i \cdot (K^h_{AP})_j}{\sqrt{d_{head}}}$. The attention logit $l^h_{i \leftarrow j}$ is computed by summing the RP bias $b^h_{i \leftarrow j}$, the semantic attention weight $w^{h,sem}_{i \leftarrow j}$, and the AP attention weight $w^{h,AP}_{i \leftarrow j}$ (the latter only when $i,j \notin \mathscr{S}$):

\begin{equation}
\label{eq:attention_score}
    l^h_{i \leftarrow j} = b^h_{i \leftarrow j} + w^{h,sem}_{i \leftarrow j}
    + w^{h,AP}_{i \leftarrow j}\mathbbm{1}[i \notin \mathscr{S} \wedge  j \notin \mathscr{S}]
\end{equation}

\noindent Attention scores are obtained by applying the softmax, normalizing over all keys $j$: $A^h_{i \leftarrow j} = \frac{\exp(l^h_{i \leftarrow j})}{\sum_{j'} \exp(l^h_{i \leftarrow j'})}$. A visual explanation of this process is shown in Appendix~\ref{subapp:attention} (Figure~\ref{fig:attn_weights_sum}).

The values are computed separately for each stream and remain within their respective spaces: per head $h$, $V^h_{AP} = W_{v}^{h,AP} x_{AP}$ and $V^h_{sem} = W_{v}^{h,sem} x_{sem}$, with $W_{v}^{h,AP} \in \mathbb{R}^{(d_{AP}/n_{heads}) \times d_{AP}}$ and $W_{v}^{h,sem} \in \mathbb{R}^{(d_{sem}/n_{heads}) \times d_{sem}}$. The same attention scores $A^h$ are applied to both value streams, and the concatenated per-head outputs are processed by output projections $W_{o}^{AP} \in \mathbb{R}^{d_{AP} \times d_{AP}}$ and $W_{o}^{sem} \in \mathbb{R}^{d_{sem} \times d_{sem}}$.


\subsection{SwiGLU} 
NeoBERT, and most recent Transformer architectures, use SwiGLU \cite{shazeer-etal-2020-glu} as a drop-in replacement for the post-attention feed-forward network. SwiGLU is defined as:

\begin{equation}
    W_{down}  \left( \text{swish}(W_{gate} x) \otimes W_{up} x\right)
\end{equation}

\noindent with $x$ the token embeddings, $W_{up}, W_{gate} \in \mathbb{R}^{d_{int} \times d_{model}}$ and $W_{down} \in \mathbb{R}^{d_{model} \times d_{int}}$ with $d_{int}$ a hidden intermediate size usually defined as $d_{int}=4 \cdot d_{model}$, $\otimes$ the Hadamard product, and $\text{swish}(x) = x\cdot sigmoid(x)$.

We implement this by allowing the up-projection $W_{up}$ and the gating $W_{gate}$ to operate on the concatenated vector $x = [x_{AP}; x_{sem}] \in \mathbb{R}^{d_{model}}$, projecting to a $d_{int}=d^{AP}_{int} + d^{sem}_{int}$ dimensional space. To maintain the architectural separation, we re-separate the information during the down-projection. Denoting by $x_{AP}^{(l)}$ and $x_{sem}^{(l)}$ the AP and semantic representations after the attention block at layer $l$:
\begin{equation*}
\begin{split}
& h_{inter} = \text{swish}(W_{gate}x) \otimes (W_{up}x) \\
& x_{AP}^{(l+1)} = x_{AP}^{(l)} + W_{down}^{AP} (h_{inter}[:d^{AP}_{int}]) \\
& x_{sem}^{(l+1)} = x_{sem}^{(l)} + W_{down}^{sem} (h_{inter}[d^{AP}_{int}:])
\end{split}
\end{equation*}

\noindent where $W_{down}^{AP} \in \mathbb{R}^{d_{AP} \times d^{AP}_{int}}$ and $W_{down}^{sem} \in \mathbb{R}^{d_{sem} \times d^{sem}_{int}}$, with $d^{AP}_{int} = 4 \cdot d_{AP}$ and $d^{sem}_{int} = 4 \cdot d_{sem}$. 

The SwiGLU is the only component where semantic and positional embeddings directly interact, serving as a channel for semantic-conditioned structural updates (e.g., a ``\textbackslash n'' token triggering a segment boundary in the AP space). The dual down-projections re-segregate the information before the residual connection. This design corroborates previous work \cite{golovneva-etal-2024-contextual, zheng-etal-2024-dape} showing the benefit of conditioning positional representations on semantic content.

\subsection{Training}
\label{subsec:training}

\paragraph{Setup} Following \citet{breton-etal-2025-neobert}, we train our architecture on the English corpus of the \emph{FineWeb} dataset \citep{penedo-etal-2024-fineweb} with a Masked Language Modeling (MLM) objective. We use an $L=6$ layers architecture with $A=6$ heads per layer and a hidden size of $d_{model} = 768$. We split the embedding into $d_{AP} = 48$ and $d_{sem} = 720$, corresponding to $1/16$ of the dimensions for positional information. This ratio is motivated by the observation that positional information is inherently low-dimensional~\cite{song-etal-2024-uncovering}. The maximum positional embedding is set to 512. We train for $70$k steps on four H100 GPUs with an effective batch size of 256, totaling around 22B tokens, using an AdamW optimizer \cite{loshchilov-etal-2019-decoupled} with a 10k-step learning rate warm-up followed by cosine decay.

\paragraph{MLM Decoding} Since MLM is an intrinsically semantic task, the decoding head is applied only to the semantic part of the last hidden state ($\mathbb{R}^{d_{sem} \times d_{vocab}}$), rather than to the full $d_{model}$-dimensional representation. This is an important design choice: it frees the AP subspace from prediction pressure. Because of this, the AP component of the last layer's hidden state is unused by the MLM head and therefore receives no gradient.

\paragraph{AP Shifting} We use random position shifting \cite{kiyono-etal-2021-shape} to uniformly train all positions and to decorrelate tokens' semantic content from their position. Given a list of $n$ tokens as input $[T_0, T_1,...,T_n]$, instead of assigning them positions $[P_0, P_1,...,P_n]$, we randomly sample $k \in [0, m-n]$ where $m$ is the maximum position embedding and assign positions $[P_{k}, P_{k+1},...,P_{k+n}]$. 

\paragraph{Baselines} In the following sections, we compare our disentangled architecture (DSTG-NeoBERT, hereafter the default DSTG variant trained with the semantic-only MLM head described above) with three baseline NeoBERT architectures: RoPE-NeoBERT, with a RoPE-based positional encoding; AP-NeoBERT, with a learned AP embedding; and RP-NeoBERT, with a learned RP embedding. The three baselines have $L=6$ layers with $A=6$ heads per layer, and a hidden size $d_{model} = 720$ (matching $d_{sem}$ of DSTG-NeoBERT; the resulting parameter asymmetry is discussed in the Limitations section). These baselines are trained using the exact same experimental setup and data.

\subsection{Preliminary Experiments}
Before any analysis, we ensure that DSTG-NeoBERT does not show catastrophic failure on standard encoder tasks, and is comparable to its entangled counterparts. We evaluate the four models on three standard benchmarks: GLUE \citep{wang-etal-2018-glue}, MTEB \citep{muennighoff-etal-2023-mteb} and SQuAD \citep{rajpurkar-etal-2016-squad}. GLUE tests natural language inference and semantic similarity across 8 datasets; MTEB evaluates zero-shot dense embeddings across 56 English datasets spanning 7 task types; SQuAD benchmarks model on extractive question answering. 
Table~\ref{tab:results_benchmarks} displays the results and confirms that disentanglement does not degrade standard performance. Experimental details and per-task results are given in Appendix~\ref{app:experiments}.

\begin{table}
    \centering
    \small
    \begin{tabular}{llccc|cc}
        \toprule
         & metric & \textbf{AP} & \textbf{RP}  & \textbf{RoPE} & \textbf{DSTG}  \\
        \midrule
        \textbf{GLUE} & Avg. & \textbf{0.79} & 0.75 & \textbf{0.79} & \underline{0.78} \\
        \textbf{MTEB} & Avg. & \textbf{0.47} & 0.43 & \underline{0.46} & \underline{0.46} \\
        \textbf{SQUAD} & EM & 0.74 & 0.75 & \underline{0.76}  & \textbf{0.77} \\
          & F1 & 0.82 & \underline{0.84} & \textbf{0.86} & \textbf{0.86} \\
        \bottomrule
    \end{tabular}
    \caption{Results on the GLUE, MTEB and SQuAD benchmarks. "Avg." metric signify an average of task-dependent metrics (see Appendix~\ref{app:experiments}). "EM" means \textit{Exact Match}.}
    \label{tab:results_benchmarks}
\end{table}



\section{Analysis of the Learned Representations}
\label{sec:analysis}

DSTG-NeoBERT encodes structural information in a low-dimensional AP subspace, and its attention heads spontaneously specialize into structure-driven and semantic-driven groups. We demonstrate these properties through visualization, probing, and spectral analysis.

\paragraph{(1) The learned AP embeddings collapse into a 2D low-frequency manifold}
Despite being 48-dimensional, DSTG-NeoBERT's learned AP embedding matrix (the lookup table mapping positions to initial AP vectors) converges toward a two-dimensional low-frequency sinusoidal space. First, PCA on this embedding matrix, following \citet{wang-chen-2020-position}, shows that the learned AP embeddings collapse into a two-dimensional manifold, with the first two principal components (PCs) capturing 94.1\% of the total spatial variance. In contrast, the AP-NeoBERT baseline's learned position embeddings remain entangled, capturing only 23.4\% in their first two PCs. Second, a Type-II Discrete Cosine Transform confirms that these PCs operate at low frequencies, concentrating 98.4\% and 99.0\% of their spectral power within the first four bins. The third component (2.3\% variance) acts as a high-frequency component. Conversely, AP-NeoBERT's top PCs are dominated by high-frequency signals, holding under 0.2\% of their power in low-frequency bins. Appendix~\ref{subapp:learned_ap_embeddings} details DSTG-NeoBERT's representational simplicity. \textbf{This corroborates findings from \citet{song-etal-2024-uncovering}, and hints at the inherent simplicity of the positional information needed by Transformers when they can use RP on top of AP.}

\begin{figure}
    \centering
    \includegraphics[width=0.75\linewidth]{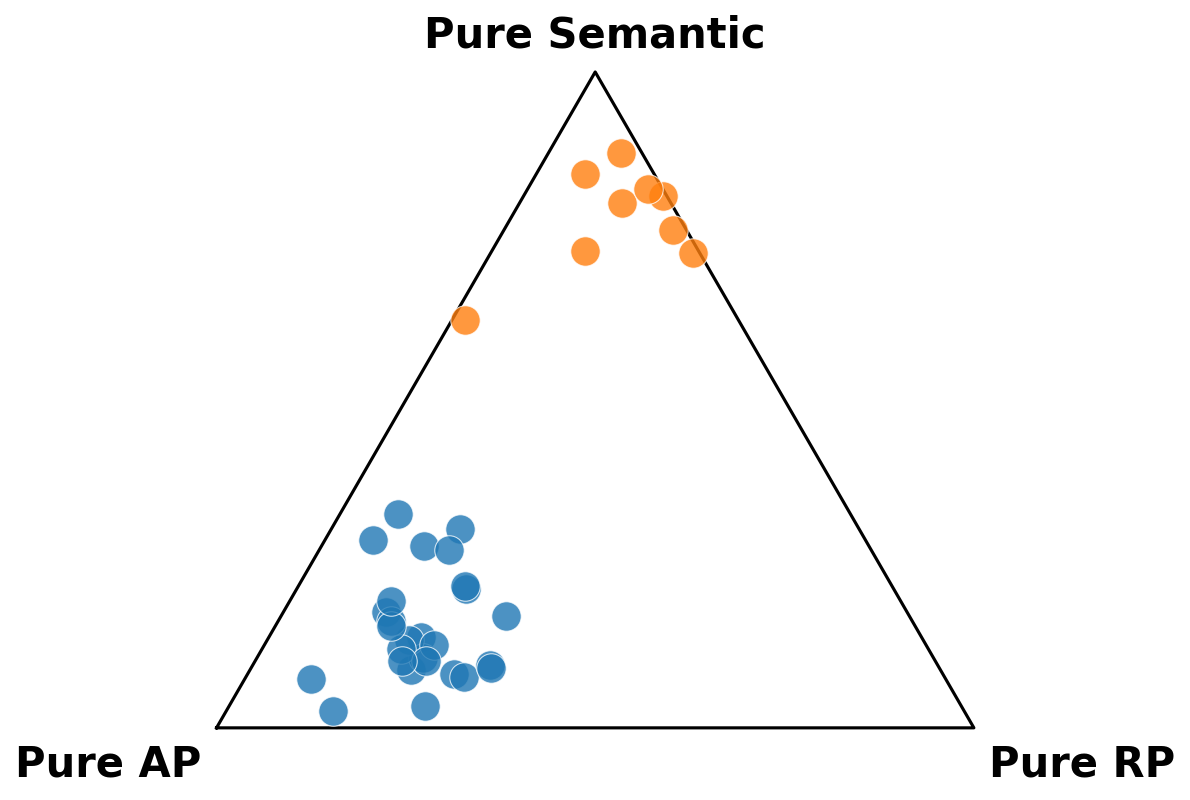}
    \caption{Categorization of all heads across layers of DSTG-NeoBERT, based on the KL-divergence of ablated information. Each corner corresponds to heads influenced by a single information.}
    \label{fig:heads_triangle}
\end{figure}

\begin{figure*}[ht]
    \centering
    \includegraphics[width=0.86\linewidth]{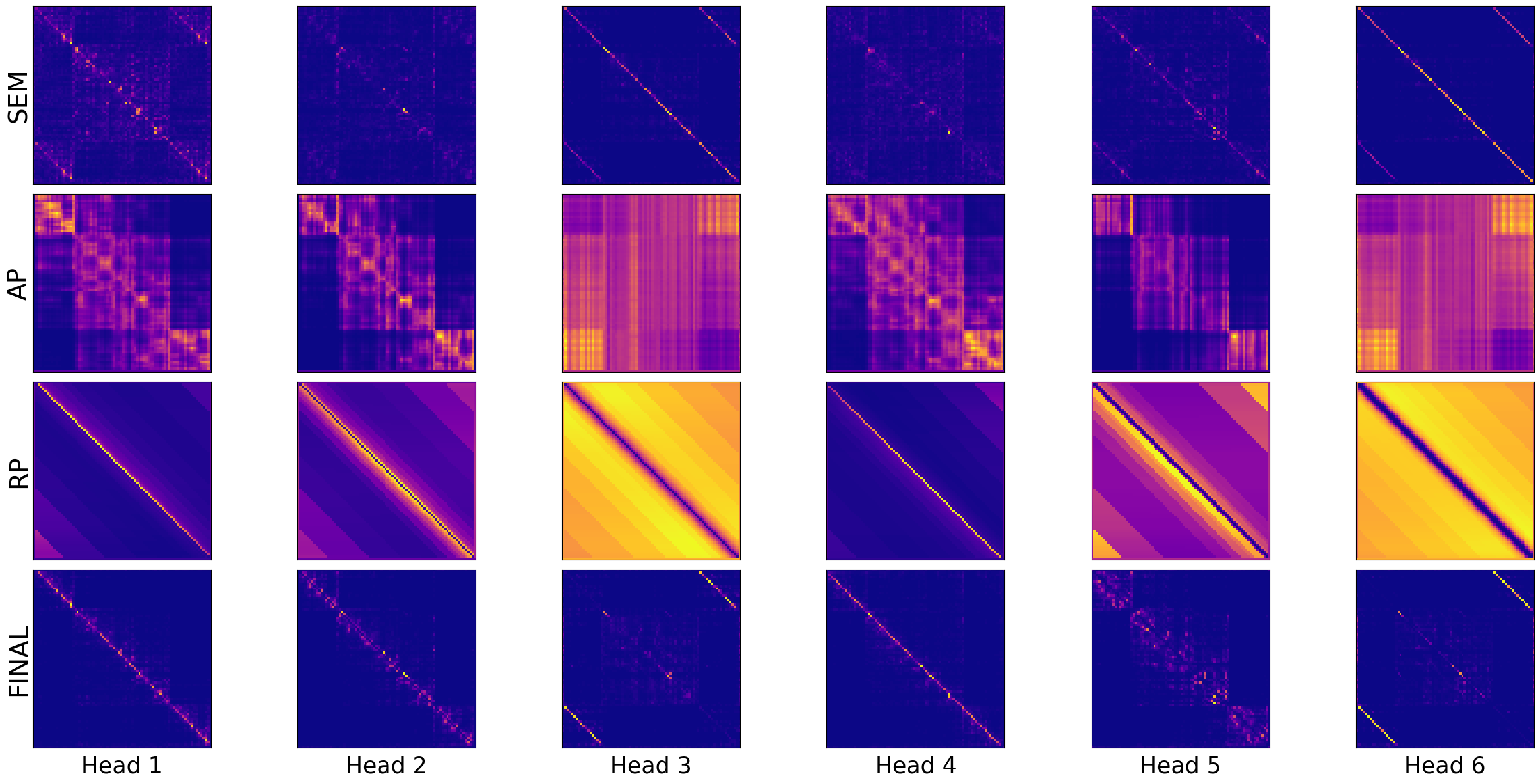}
    \caption{Softmax applied independently to the last layer's attention weights for semantic (1st row), AP (2nd row) and RP (3rd row) components, as well as the combined attention scores (Eq.~\ref{eq:attention_score}, 4th row) on an example of shape DOC-A + DOC-B + DOC-A. Each column corresponds to an attention head. Color scale differs across plots.}
    \label{fig:attention_patterns}
\end{figure*}

\begin{figure*}[ht]
    \centering
    \includegraphics[width=0.95\linewidth]{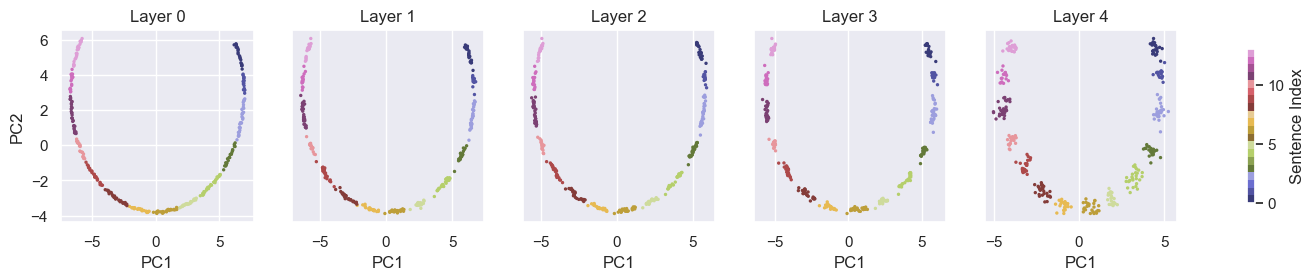}
    \caption{2-dimensional PCA of the AP hidden states at each layer when encoding a long document. Each sentence (delimited by punctuation) is given a different color. Tokens from the same sentence cluster together, with clusters becoming increasingly distinct in deeper layers. The last layer is omitted because the AP subspace is not used for MLM prediction, leaving it unconstrained.}
    \label{fig:AP_hidden_state_PCA}
\end{figure*}

\paragraph{(2) DSTG-NeoBERT trades off absolute position for structure}
The model uses its low-frequency AP components to represent the macroscopic structure of the text. As exhibited in its attention patterns (Figure~\ref{fig:attention_patterns}, second row), DSTG-NeoBERT learns to use the AP embedding space as structural encoding, separating sentences and paragraphs into blocks. Additionally, visualizing the AP \emph{hidden states} across layers (Figure~\ref{fig:AP_hidden_state_PCA}) -- whose first two PCs retain about 90\% of the variance -- reveals a growing separation between sentences across layers: tokens from the same sentence cluster together, and these clusters become increasingly distinct in deeper layers. The AP space therefore keeps its low effective dimensionality throughout the network. \textbf{This shows that the AP space progressively abstracts raw position into document structure.}

\paragraph{(3) The post-attention SwiGLU is the component allowing for structural encoding}
We observed that the AP stream learns to encode document structure, and found that the post-attention SwiGLU is a key component in this process. We trained another DSTG-NeoBERT in which the semantic-AP mixing SwiGLU is separated into two distinct SwiGLUs: one for the semantic embeddings and one for the AP embeddings. In this setting, the model is unable to encode structure in its AP space, and we do not observe the structural learning shown in Figures~\ref{fig:attention_patterns} and~\ref{fig:AP_hidden_state_PCA}. This also highlights that the shared attention scores alone seem insufficient to learn structural encoding. \textbf{The SwiGLU component, in which semantic and positional information can interact directly, is a key component for the model to understand document structure.}
\paragraph{(4) Attention heads cluster into AP-oriented and semantic-oriented groups}
DSTG-NeoBERT’s disentangled architecture reveals a functional taxonomy: attention heads specialize almost exclusively as either AP-driven or semantic-driven. Notably, we find no purely RP-oriented heads. Instead, the RP bias acts as a localized support mechanism for the semantic cluster.
As shown in Figure~\ref{fig:attention_patterns}, heads 3 and 6 act as \emph{retrieval} heads: The semantic attention (first row) attends to semantically similar tokens, while the RP bias (third row) disallows tokens from attending to themselves, enabling the model to focus entirely on distant context.\newline
We empirically establish this taxonomy by quantifying the weight of each representational space through information ablation. We compute KL-divergence between the true attention probability distribution ($A^h$) and the ablated distribution ($A^h_{\setminus i}$) generated when a component $i$ (AP, RP or semantic) is removed prior to the softmax, for each information $i \in \{sem, AP, RP\}$:

\begin{equation}
    Score_{i} = D_{KL}(A^h \| A^h_{\setminus i})
\end{equation}

 Averaging these KL divergences across 500 documents from WikiText \citep{merity2016pointer} yields a 3-dimensional influence vector $[Score_{sem}, Score_{AP}, Score_{RP}]$ for each head, normalized per-head. 
 As shown in Figure~\ref{fig:heads_triangle}, only 9 out of the 36 heads \textit{specialize} in semantic matching. The remaining heads are strongly affected by the AP information. \textbf{Despite being given the freedom to either use AP or RP positional information, the model only uses RP as complementary information to the semantic heads, while AP is used as its own information stream.}
\newline
\newline
These findings suggest that while RP and semantic information complement each other, AP information is used as an additional structural information stream. This raises two questions: (1) Do entangled models with other positional encodings (RP, RoPE, or entangled AP) end up encoding comparable structural information in their hidden states, or is the disentangled AP stream playing a distinct role? (2) Does the structural information preserved by the disentangled architecture translate into gains on downstream linguistic probes? We answer these questions in the next section.


\section{Probing Structural Representation in Transformers}
\label{sec:AP_necessary_struct}
Understanding document structure requires a hierarchical coordinate system capturing a token's global absolute position, its structural segment, and its local intra-segment position. This is what we study in this section.

Let model $\mathcal{M}$ map a document $\mathcal{D}$  of length $N$ to hidden states $H^{(l)} = \{h_1^{(l)}, \dots, h_N^{(l)}\}$ at layer $l$. We evaluate the four models using three linear probes, one per hierarchy level, defined as $f(h_i^{(l)}) = W^\top h_i^{(l)} + b$. Probes are optimized via Ridge regression and evaluated using $R^2$. We use texts from the WikiText \citep{merity2016pointer} dataset, concatenated to make 500 documents of maximum input lengths for the models (512 tokens). We report the average results over five runs with different seeds, using 400 documents for training and 100 for testing. For DSTG-NeoBERT, we also probe separately the AP and semantic spaces to evaluate where the information is represented.

\subsection{Token-level AP}
\label{subsec:token_ap}
The first level of the hierarchy is the most basic: can a model recover the global position of a token? 
The absolute position $y_i \in [0, 1]$ of token $t_i$ is defined as its normalized index:
$$
y_i = \frac{pos(t_i)}{max_j(pos(t_j))} =  \frac{pos(t_i)}{512}
$$


\paragraph{Results} The results are reported in Table~\ref{tab:token-AP-probe}. As expected, AP-NeoBERT properly encodes this information, especially in early layers. Critically, AP information collapses in the final layer ($R^2 = 0.34$): we empirically show that this behavior is at least partly due to the MLM training objective in Section~\ref{subsec:mlm}. 
RoPE-NeoBERT and RP-NeoBERT are largely unable to encode AP information in their hidden states.

\begin{table}[h]
    \centering
    \small
    \begin{tabular}{lcccccc}
        \toprule
        \textbf{Layer} & \textbf{0} & \textbf{1} & \textbf{2} & \textbf{3} & \textbf{4} & \textbf{5} \\
        \midrule
        \textbf{RoPE} & 0.10 & 0.12 & 0.32 & 0.20 & 0.18 & 0.06 \\
        
        \textbf{RP} & 0.17 & 0.50 & 0.46 & 0.42 & 0.42 & 0.22 \\
        
        \textbf{AP} & \textbf{0.99} & \textbf{0.99} & \textbf{0.97} & \underline{0.89} & \underline{0.82} & 0.24 \\
        \midrule
        \textbf{DSTG} & \textbf{0.99} & \underline{0.98} & \underline{0.96} & \textbf{0.95} & \textbf{0.92} & 
        - \\ 
        
        \ \ \footnotesize \textit{AP only} & \footnotesize \textit{0.99} & \footnotesize \textit{0.98} & \footnotesize \textit{0.95} & \footnotesize \textit{0.94} & \footnotesize \textit{0.89} & \footnotesize - \\ 
        \ \ \footnotesize \textit{sem. only} & \footnotesize \textit{0.44} & \footnotesize \textit{0.34} & \footnotesize \textit{0.35} & \footnotesize \textit{0.32} & \footnotesize \textit{0.26} & \footnotesize \textit{0.18} \\
        \bottomrule
    \end{tabular}
    \caption{Token-level AP probe results. ``-'' marks the last-layer AP probe for DSTG: because the AP component of the final hidden state receives no gradient (see Section~\ref{subsec:training}), we do not probe it; we report only the semantic subspace at layer 5.}
    \label{tab:token-AP-probe}
\end{table}

\subsection{Segment-level AP} 
\label{subsec:segment_ap}
The second level tests coarse structural membership: does a model know which segment (e.g., sentence or paragraph) a token belongs to?
Let a document $\mathcal{D}$ be partitioned into a sequence of $K$ segments $\mathcal{S} = \{S_0, S_1, \dots, S_{K-1}\}$. We define a mapping function $\sigma(t_i) \to \{0, \dots, K-1\}$ that assigns each token $t_i$ to its segment index. The segment index is updated at every structural boundary (defined by terminal punctuation or newline characters). Unlike the absolute position probe, the target $y_i$ for this task is the discrete segment ID:$$y_i = \sigma(t_i)$$This target measures the model's awareness of its current progress through the document's macroscopic structure, independent of the token's local position within a specific segment.

\paragraph{Results} As displayed in Table~\ref{tab:segment-AP-probe}, the results follow the same pattern as the token-level probe. AP-NeoBERT strongly encodes segment-level AP in early layers but loses the information in the last layer, while DSTG-NeoBERT manages to keep this structural information. Both relative baselines struggle to encode segment-level position. 

\begin{table}[h]
    \centering
    \small
    \begin{tabular}{lcccccc}
        \toprule
        \textbf{Layer} & \textbf{0} & \textbf{1} & \textbf{2} & \textbf{3} & \textbf{4} & \textbf{5} \\
        \midrule
        \textbf{RoPE} & 0.09 & 0.11 & 0.30 & 0.19 & 0.17 & 0.06 \\
        \textbf{RP} & 0.17 & 0.47 & 0.40 & 0.37 & 0.37 & 0.23 \\
        \textbf{AP} & \underline{0.94} & \textbf{0.95} & \textbf{0.93} & \underline{0.85} & \underline{0.79} & 0.23 \\
        \midrule
        \textbf{DSTG} & \textbf{0.94} & \underline{0.94} & \underline{0.92} & \textbf{0.91} & \textbf{0.88} & 
         - \\ 
        \ \ \footnotesize \textit{AP only} & \footnotesize \textit{0.94} & \footnotesize \textit{0.93} & \footnotesize \textit{0.91} & \footnotesize \textit{0.90} & \footnotesize \textit{0.85} & \footnotesize - \\ 
        \ \ \footnotesize \textit{sem. only} & \footnotesize \textit{0.42} & \footnotesize \textit{0.33} & \footnotesize \textit{0.35} & \footnotesize \textit{0.31} & \footnotesize \textit{0.27} & \footnotesize \textit{0.18} \\
        \bottomrule
    \end{tabular}
    \caption{Segment-level AP probe results. ``-'' marks the discarded last AP space (see Table~\ref{tab:token-AP-probe} caption).}
    \label{tab:segment-AP-probe}
\end{table}

\subsection{Intra-segment token AP}
\label{subsec:intrasegment_ap}
The third level tests local progress: does a model encode where a token sits within its own segment?
For a token $t_i$ in segment $S_k$ of length $L_k$, we define the target variable $y_i^{rel} \in [0, 1]$ as the normalized progress within that segment:$$y_i^{rel} = \frac{pos(t_i) - \min \{j \mid t_j \in S_k\}}{L_k - 1}$$where $pos(t_i)$ is the absolute index of the token. This target transforms a discrete sawtooth count into a linear slope, where $0.0$ denotes the segment start and $1.0$ the terminal boundary marker.

\paragraph{Results}
Though AP-NeoBERT and DSTG-NeoBERT perform better than RoPE- and RP- NeoBERT, all four perform correctly on the task, as displayed in Table~\ref{tab:intrasegment-AP-probe}.  This unexpected result is explained in the subspace probing of DSTG-NeoBERT: we find that the entirety of the intra-segment position information can be retrieved from the semantic subspace, and that the positional subspace plays no role in this probe.

\begin{table}[h]
    \centering
    \small
    \begin{tabular}{lcccccc}
        \toprule
        \textbf{Layer} & \textbf{0} & \textbf{1} & \textbf{2} & \textbf{3} & \textbf{4} & \textbf{5} \\
        \midrule
        \textbf{RoPE} & 0.29 & 0.70 & 0.73 & 0.69 & \underline{0.68} & \underline{0.50} \\
        
        \textbf{RP} & \textbf{0.50} & 0.68 & 0.67 & 0.68 & 0.63 & \underline{0.50} \\
        \textbf{AP} & 0.20 & \textbf{0.76} & \textbf{0.80} &\textbf{0.76} & 0.66 & 0.46 \\

        \midrule
        
        \textbf{DSTG} & \underline{0.43} & \underline{0.73} & \underline{0.76} & \textbf{0.76} & \textbf{0.73} & 
        - \\ 

        \ \ \footnotesize \textit{AP only} & \footnotesize \textit{0.04} & \footnotesize \textit{0.06} & \footnotesize \textit{0.06} & \footnotesize \textit{0.06} & \footnotesize \textit{0.06} & \footnotesize - \\ 
        \ \ \footnotesize \textit{sem. only} & \footnotesize \textit{0.42} & \footnotesize \textit{0.73} & \footnotesize \textit{0.76} & \footnotesize \textit{0.75} & \footnotesize \textit{0.73} & \footnotesize \textit{0.58} \\
        \bottomrule
    \end{tabular}
    \caption{Intra-segment token AP probe results. ``-'' marks the discarded last AP subspace (Table~\ref{tab:token-AP-probe} caption).}
    \label{tab:intrasegment-AP-probe}
\end{table}

\subsection{Effect of the MLM training objective on positional representations}
\label{subsec:mlm}
The three probes displayed similar patterns on the baselines: the positional information is hardly linearly decodable in the last layer, even when encoded up to the second-to-last layer. 
We hypothesize that because MLM is a semantic task and that the MLM head operates on the full hidden state, the model discard positional information for semantic prediction. 
\newline
We tested this hypothesis through an additional experiment: we trained another instance of DSTG-NeoBERT with the same training setup as described in Section~\ref{subsec:training}, except that
the MLM head is applied on the complete $d_{model}=d_{AP} + d_{sem}$ representations rather than only on $d_{sem}$. We then run the token-level and segment-level AP probes on this model, and compare the results to the standard DSTG-NeoBERT at each layer. The results are shown in Appendix~\ref{subapp:intermodel} (Figure~\ref{fig:DSTG_probes}): while the probes results are similar for layers 0 to 4, only the model trained with the MLM head on both positional and semantic information displays a significant decrease in $R^2$ when probing its positional representations of the last-layer. This experiment isolated the effect of MLM on the positional representations. 

Our experiments establish two findings: 
(1) \textbf{among the encoders we study, only those that carry explicit and persistent AP information in their hidden states reliably encode document structure}; relative-only models (RP, RoPE) do not learn to do so under MLM pretraining; and (2) \textbf{AP-NeoBERT does encode structure and uses it in its intermediate layers, but discards this information in its last layer} to prepare for the MLM task.


\subsection{Benefits of the Structural Encoding}
\label{sec:experiments}
\todo[inline]{relire}

We hypothesize that the additional structural information enhances token representations. To validate this, we run the disentangled model and the three baselines on the Flash-Holmes \citep{waldisetal2024holmes} probing benchmark. Flash-Holmes uses over 200 datasets to probe hidden representations on 65 linguistic phenomena making up 5 fields: \textit{Morphology} probes the structure of words with phenomena such as irregular forms or subject-verb agreement; \textit{Syntax} probes the structure of sentences such as filler gaps or argument structure; \textit{Discourse} probes the context in text such as rhetorical structure or sentence order; \textit{Semantics} probes the meaning of words such as subject gender or factuality; and \textit{Reasoning} probes the use of words in logical deduction such as negation.\footnote{Definitions taken directly from \citet{waldisetal2024holmes}.} \newline
We provide the results for the AP and RP baselines with $d_{model}=720$, and DSTG-NeoBERT with $d_{sem}=720$ and $d_{pos}=48$. For RoPE-NeoBERT, we evaluate two models with $d_{model}=720$ and $d_{model}=768$. The former is the baseline used in the previous section with the same semantic representation size, while the latter is used as a matched-parameters baseline to account for the model size difference with DSTG-NeoBERT. The results are averaged over five runs with different seeds and reported in Table~\ref{tab:holmes-results}. DSTG-NeoBERT performs best in all fields but Morphology, for which it arrives second to the matched-parameters RoPE baseline. Interestingly, the three $d_{model}=720$ baselines perform around equally and DSTG still beats the $d_{model}=768$ RoPE baseline on four of 5 fields. This hints that the additional gains stem from the preserved structural information. We report the results on the 65 phenomena in Appendix~\ref{subapp:holmes}. 
The disentangled architecture obtains the best score on 42 of the 65 phenomena (including 30 with no ties): 14 out of 21 phenomena in \textit{Syntax}, 16 out of 24 in \textit{Semantics}, 5 out of 6 in \textit{Discourse}, 6 out of 10 in \textit{Reasoning}, and 1 out of 4 in \textit{Morphology}.


\begin{table}
    \centering
    \small
    \begin{tabular}{lcccc|c}
        \toprule
        \textbf{Model} & \textbf{AP} & \textbf{RP} & \multicolumn{2}{c|}{\textbf{RoPE}} & \textbf{DSTG} \\
        \cmidrule(lr){2-2} \cmidrule(lr){3-3} \cmidrule(lr){4-5} \cmidrule(lr){6-6}
        Hidden Size & 720 & 720 &  720 &  768 & $720+48$\\
        \midrule
        \textbf{Morphology} & 0.58 & 0.58 & 0.58 & \textbf{0.63} & 0.60 \\ 
        \textbf{Reasoning}  & 0.47 & 0.47 & 0.46 & 0.47 & \textbf{0.48} \\ 
        \textbf{Semantics}  & 0.43 & 0.43 & 0.45 & 0.43 & \textbf{0.46} \\ 
        \textbf{Discourse}  & 0.32 & 0.33 & 0.32 & 0.32 & \textbf{0.35} \\ 
        \textbf{Syntax}     & 0.71 & 0.71 & 0.71 & 0.73 & \textbf{0.75} \\      
        \bottomrule
    \end{tabular}
    \caption{Results on the five linguistic fields of the Flash-Holmes benchmark \citep{waldisetal2024holmes}. \textit{Hidden Size} refers to $d_{model}$ for AP, RP and RoPE baselines, and $d_{sem} + d_{pos}$ for DSTG.}
    \label{tab:holmes-results}
\end{table}

\section{Conclusion}
\todo[inline]{relire}
In this work, we explored the explicit disentanglement of semantic, absolute positional, and relative positional information in Transformer encoders. By introducing dedicated positional and semantic streams, we showed that positional information naturally organizes into a simple low-dimensional structural manifold, while attention heads specialize into distinct structural and semantic roles. Our analysis further revealed that standard positional encoding methods either fail to preserve macroscopic structure (for relative-bias-based methods), or discard it in later layers under MLM training pressure (for AP methods). Removing AP information from the MLM objective allowed the model to preserve structural information throughout the network and improved linguistic representations on probing tasks, outperforming entangled baselines on most phenomena in the Flash-Holmes benchmark and matching them on GLUE, MTEB and SQuAD. \newline
Our work suggests new approaches to positional encoding in transformers. The simplicity of the learned disentangled AP space, paired with our findings on the loss of positional information in the last layer due to MLM, hints at the creation of new explicit ways to provide the model with absolute positional information without interfering with semantic representations. Additionally, if our observations extend to decoders, a disentangled approach opens practical avenues: caching position-independent semantic representations for RAG, leveraging head specialization for selective computation, and manipulating the low-dimensional AP manifold for length extrapolation. Therefore, scaling to larger models and extending the approach to decoder-only architectures are promising directions for future work.

\section*{Limitations}
While our work successfully demonstrates the theoretical and empirical benefits of disentangling positional and semantic information, this exploratory work has several notable limitations. Our empirical validation is currently constrained to a relatively small scale. The disentangled model and baselines were trained from scratch on approximately 22 billion tokens using a 6-layer, 6-head architecture. While this scale is sufficient to show the emergence of the low-frequency structural manifold and the functional taxonomy of attention heads, we have not yet verified whether these disentanglement properties hold, or whether the training dynamics shift, when scaled to massive parameter counts (e.g., 7B+ parameters) or larger token budgets.

Additionally, DSTG-NeoBERT uses a slightly larger hidden size than the three baselines ($d_{model}=768$ vs. $720$), which gives it $\sim 6\%$ more parameters per token. We made this choice so that the semantic stream of DSTG-NeoBERT matches the full hidden size of the baselines, but it introduces a small capacity asymmetry in the comparisons. The maximum sequence length is also limited to 512 tokens, which restricts our ability to probe the very long-context regimes that motivated this work; we expect the AP manifold's low-frequency structure to be most useful at longer contexts, but verifying this requires a separate scaling study.

Furthermore, our architectural modifications and probing experiments are specifically designed for and evaluated on encoder-only Transformers. We intentionally focused on bidirectional models because they cannot implicitly infer position through a causal attention mask, ensuring a cleaner separation of variables for our analysis. Consequently, the applicability of this strict three-stream separation to decoder-only architectures remains an open question for future work.

\section*{Acknowledgments}
This research was funded by BPI-France under the project AI For Democracy - Democratic Commons, one of seven winners of BPI-France’s “Digital Commons for Generative AI” call for projects, conducted as part of the France 2030 investment plan. We thank members of the Democratic Commons project, as well as Lise Le Boudec for the careful proof reading. This work was performed using HPC resources from GENCI–IDRIS (Grant 2025-AD011015927R1).

\bibliography{custom, anthology-1, anthology-2}

\begin{thebibliography}{40}
\providecommand{\natexlab}[1]{#1}

\bibitem[{Chen et~al.(2025)Chen, Lv, Luan, Wang, and Liu}]{chen-etal-2025-hope}
Yuhan Chen, Ang Lv, Jian Luan, Bin Wang, and Wei Liu. 2025.
\newblock \href {https://doi.org/10.18653/v1/2025.acl-long.1123} {{H}o{PE}: A novel positional encoding without long-term decay for enhanced context awareness and extrapolation}.
\newblock In \emph{Proceedings of the 63rd Annual Meeting of the Association for Computational Linguistics (Volume 1: Long Papers)}, pages 23044--23056, Vienna, Austria. Association for Computational Linguistics.

\bibitem[{Chi et~al.(2022)Chi, Fan, Ramadge, and Rudnicky}]{chi-etal-2022-kerple}
Ta-Chung Chi, Ting-Han Fan, Peter~J Ramadge, and Alexander Rudnicky. 2022.
\newblock Kerple: Kernelized relative positional embedding for length extrapolation.
\newblock \emph{Advances in Neural Information Processing Systems}, 35:8386--8399.

\bibitem[{Chi et~al.(2023)Chi, Fan, Rudnicky, and Ramadge}]{chi-etal-2023-dissecting}
Ta-Chung Chi, Ting-Han Fan, Alexander Rudnicky, and Peter Ramadge. 2023.
\newblock \href {https://doi.org/10.18653/v1/2023.acl-long.756} {Dissecting transformer length extrapolation via the lens of receptive field analysis}.
\newblock In \emph{Proceedings of the 61st Annual Meeting of the Association for Computational Linguistics (Volume 1: Long Papers)}, pages 13522--13537, Toronto, Canada. Association for Computational Linguistics.

\bibitem[{Clark et~al.(2020)Clark, Luong, Le, and Manning}]{clark-etal-2020-electra}
Kevin Clark, Minh-Thang Luong, Quoc~V. Le, and Christopher~D. Manning. 2020.
\newblock \href {https://doi.org/10.48550/arXiv.2003.10555} {{{ELECTRA}}: {{Pre-training Text Encoders}} as {{Discriminators Rather Than Generators}}}.
\newblock \emph{Preprint}, arXiv:2003.10555.

\bibitem[{Devlin et~al.(2019)Devlin, Chang, Lee, and Toutanova}]{devlin-etal-2019-bert}
Jacob Devlin, Ming-Wei Chang, Kenton Lee, and Kristina Toutanova. 2019.
\newblock \href {https://doi.org/10.18653/v1/N19-1423} {{BERT}: Pre-training of deep bidirectional transformers for language understanding}.
\newblock In \emph{Proceedings of the 2019 Conference of the North {A}merican Chapter of the Association for Computational Linguistics: Human Language Technologies, Volume 1 (Long and Short Papers)}, pages 4171--4186, Minneapolis, Minnesota. Association for Computational Linguistics.

\bibitem[{Flesch(1948)}]{flesch1948new}
Rudolph Flesch. 1948.
\newblock A new readability yardstick.
\newblock \emph{Journal of applied psychology}, 32(3):221.

\bibitem[{Golovneva et~al.(2024)Golovneva, Wang, Weston, and Sukhbaatar}]{golovneva-etal-2024-contextual}
Olga Golovneva, Tianlu Wang, Jason Weston, and Sainbayar Sukhbaatar. 2024.
\newblock \href {https://doi.org/10.48550/arXiv.2405.18719} {Contextual {{Position Encoding}}: {{Learning}} to {{Count What}}'s {{Important}}}.
\newblock \emph{Preprint}, arXiv:2405.18719.

\bibitem[{Gopalakrishnan et~al.(2026)Gopalakrishnan, Csord{\'a}s, Schmidhuber, and Mozer}]{gopalakrishnan-etal-2026-decoupling}
Anand Gopalakrishnan, Robert Csord{\'a}s, J{\"u}rgen Schmidhuber, and Michael~C. Mozer. 2026.
\newblock \href {https://doi.org/10.48550/arXiv.2509.10534} {Decoupling the "{{What}}" and "{{Where}}" {{With Polar Coordinate Positional Embeddings}}}.
\newblock \emph{Preprint}, arXiv:2509.10534.

\bibitem[{Gu et~al.(2026)Gu, Chen, Zhang, Zhang, and Hu}]{gu2026deconstructing}
Zihan Gu, Ruoyu Chen, Han Zhang, Hua Zhang, and Yue Hu. 2026.
\newblock \href {https://openreview.net/forum?id=D0u0glT060} {Deconstructing positional information: From attention logits to training biases}.
\newblock In \emph{The Fourteenth International Conference on Learning Representations}.

\bibitem[{He et~al.(2024)He, Feng, Luo, Yang, Wang, Xu, Zhang, Yang, and He}]{he-etal-2024-two}
Zhenyu He, Guhao Feng, Shengjie Luo, Kai Yang, Liwei Wang, Jingjing Xu, Zhi Zhang, Hongxia Yang, and Di~He. 2024.
\newblock \href {https://doi.org/10.48550/arXiv.2401.16421} {Two {{Stones Hit One Bird}}: {{Bilevel Positional Encoding}} for {{Better Length Extrapolation}}}.
\newblock \emph{Preprint}, arXiv:2401.16421.

\bibitem[{Ke et~al.(2021)Ke, He, and Liu}]{ke-etal-2021-rethinking}
Guolin Ke, Di~He, and Tie-Yan Liu. 2021.
\newblock \href {https://doi.org/10.48550/arXiv.2006.15595} {Rethinking {{Positional Encoding}} in {{Language Pre-training}}}.
\newblock \emph{Preprint}, arXiv:2006.15595.

\bibitem[{Kiyono et~al.(2021)Kiyono, Kobayashi, Suzuki, and Inui}]{kiyono-etal-2021-shape}
Shun Kiyono, Sosuke Kobayashi, Jun Suzuki, and Kentaro Inui. 2021.
\newblock \href {https://doi.org/10.18653/v1/2021.emnlp-main.266} {{SHAPE}: {S}hifted absolute position embedding for transformers}.
\newblock In \emph{Proceedings of the 2021 Conference on Empirical Methods in Natural Language Processing}, pages 3309--3321, Online and Punta Cana, Dominican Republic. Association for Computational Linguistics.

\bibitem[{Lan et~al.(2020)Lan, Chen, Goodman, Gimpel, Sharma, and Soricut}]{lan-etal-2020-albert}
Zhenzhong Lan, Mingda Chen, Sebastian Goodman, Kevin Gimpel, Piyush Sharma, and Radu Soricut. 2020.
\newblock \href {https://doi.org/10.48550/arXiv.1909.11942} {{{ALBERT}}: {{A Lite BERT}} for {{Self-supervised Learning}} of {{Language Representations}}}.
\newblock \emph{Preprint}, arXiv:1909.11942.

\bibitem[{Le~Breton et~al.(2025)Le~Breton, Fournier, Morris, El~Mezouar, and Chandar}]{breton-etal-2025-neobert}
Lola Le~Breton, Quentin Fournier, John~Xavier Morris, Mariam El~Mezouar, and Sarath Chandar. 2025.
\newblock \href {https://mlanthology.org/tmlr/2025/breton2025tmlr-neobert/} {{NeoBERT: A Next Generation BERT}}.
\newblock \emph{Transactions on Machine Learning Research}.

\bibitem[{Li et~al.(2026)Li, Zhao, Cai, and Sproat}]{li-etal-2026-repo}
Huayang Li, Tianyu Zhao, Deng Cai, and Richard Sproat. 2026.
\newblock \href {https://doi.org/10.48550/arXiv.2512.14391} {{{RePo}}: {{Language Models}} with {{Context Re-Positioning}}}.
\newblock \emph{Preprint}, arXiv:2512.14391.

\bibitem[{Li et~al.(2024)Li, You, Guruganesh, Ainslie, Ontanon, Zaheer, Sanghai, Yang, Kumar, and Bhojanapalli}]{li-etal-2024-functional}
Shanda Li, Chong You, Guru Guruganesh, Joshua Ainslie, Santiago Ontanon, Manzil Zaheer, Sumit Sanghai, Yiming Yang, Sanjiv Kumar, and Srinadh Bhojanapalli. 2024.
\newblock \href {https://doi.org/10.48550/arXiv.2310.04418} {Functional {{Interpolation}} for {{Relative Positions Improves Long Context Transformers}}}.
\newblock \emph{Preprint}, arXiv:2310.04418.

\bibitem[{Liu et~al.(2020)Liu, Yu, Dhillon, and Hsieh}]{liu-etal-2020-learningencode}
Xuanqing Liu, Hsiang-Fu Yu, Inderjit Dhillon, and Cho-Jui Hsieh. 2020.
\newblock \href {https://proceedings.mlr.press/v119/liu20n.html} {Learning to encode position for transformer with continuous dynamical model}.
\newblock In \emph{Proceedings of the 37th International Conference on Machine Learning}, volume 119 of \emph{Proceedings of Machine Learning Research}, pages 6327--6335. PMLR.

\bibitem[{Liu et~al.(2019)Liu, Ott, Goyal, Du, Joshi, Chen, Levy, Lewis, Zettlemoyer, and Stoyanov}]{liu-etal-2019-roberta}
Yinhan Liu, Myle Ott, Naman Goyal, Jingfei Du, Mandar Joshi, Danqi Chen, Omer Levy, Mike Lewis, Luke Zettlemoyer, and Veselin Stoyanov. 2019.
\newblock \href {https://doi.org/10.48550/arXiv.1907.11692} {{{RoBERTa}}: {{A Robustly Optimized BERT Pretraining Approach}}}.
\newblock \emph{Preprint}, arXiv:1907.11692.

\bibitem[{Loshchilov and Hutter(2019)}]{loshchilov-etal-2019-decoupled}
Ilya Loshchilov and Frank Hutter. 2019.
\newblock \href {https://doi.org/10.48550/arXiv.1711.05101} {Decoupled {{Weight Decay Regularization}}}.
\newblock \emph{Preprint}, arXiv:1711.05101.

\bibitem[{Merity et~al.(2016)Merity, Xiong, Bradbury, and Socher}]{merity2016pointer}
Stephen Merity, Caiming Xiong, James Bradbury, and Richard Socher. 2016.
\newblock \href {https://arxiv.org/abs/1609.07843} {Pointer sentinel mixture models}.
\newblock \emph{Preprint}, arXiv:1609.07843.

\bibitem[{Muennighoff et~al.(2023)Muennighoff, Tazi, Magne, and Reimers}]{muennighoff-etal-2023-mteb}
Niklas Muennighoff, Nouamane Tazi, Loic Magne, and Nils Reimers. 2023.
\newblock \href {https://doi.org/10.18653/v1/2023.eacl-main.148} {{MTEB}: Massive text embedding benchmark}.
\newblock In \emph{Proceedings of the 17th Conference of the European Chapter of the Association for Computational Linguistics}, pages 2014--2037, Dubrovnik, Croatia. Association for Computational Linguistics.

\bibitem[{Penedo et~al.(2024)Penedo, Kydl{\'\i}{\v c}ek, {allal}, Lozhkov, Mitchell, Raffel, Von~Werra, and Wolf}]{penedo-etal-2024-fineweb}
Guilherme Penedo, Hynek Kydl{\'\i}{\v c}ek, Loubna~Ben {allal}, Anton Lozhkov, Margaret Mitchell, Colin Raffel, Leandro Von~Werra, and Thomas Wolf. 2024.
\newblock \href {https://doi.org/10.52202/079017-0970} {The {{FineWeb}} datasets: {{Decanting}} the web for the finest text data at scale}.
\newblock In \emph{Advances in Neural Information Processing Systems}, volume~37, pages 30811--30849. Curran Associates, Inc.

\bibitem[{Peng et~al.(2023)Peng, Quesnelle, Fan, and Shippole}]{peng-etal-2023-yarn}
Bowen Peng, Jeffrey Quesnelle, Honglu Fan, and Enrico Shippole. 2023.
\newblock \href {https://doi.org/10.48550/arXiv.2309.00071} {{{YaRN}}: {{Efficient Context Window Extension}} of {{Large Language Models}}}.
\newblock \emph{Preprint}, arXiv:2309.00071.

\bibitem[{Press et~al.(2022)Press, Smith, and Lewis}]{press-etal-2022-train}
Ofir Press, Noah~A. Smith, and Mike Lewis. 2022.
\newblock \href {https://doi.org/10.48550/arXiv.2108.12409} {Train {{Short}}, {{Test Long}}: {{Attention}} with {{Linear Biases Enables Input Length Extrapolation}}}.
\newblock \emph{Preprint}, arXiv:2108.12409.

\bibitem[{Raffel et~al.(2020)Raffel, Shazeer, Roberts, Lee, Narang, Matena, Zhou, Li, and Liu}]{raffel-etal-2020-exploring}
Colin Raffel, Noam Shazeer, Adam Roberts, Katherine Lee, Sharan Narang, Michael Matena, Yanqi Zhou, Wei Li, and Peter~J. Liu. 2020.
\newblock Exploring the limits of transfer learning with a unified text-to-text transformer.
\newblock \emph{Journal of Machine Learning Research}, 21(140):1--67.

\bibitem[{Rajpurkar et~al.(2016)Rajpurkar, Zhang, Lopyrev, and Liang}]{rajpurkar-etal-2016-squad}
Pranav Rajpurkar, Jian Zhang, Konstantin Lopyrev, and Percy Liang. 2016.
\newblock \href {https://doi.org/10.18653/v1/D16-1264} {{SQ}u{AD}: 100,000+ questions for machine comprehension of text}.
\newblock In \emph{Proceedings of the 2016 Conference on Empirical Methods in Natural Language Processing}, pages 2383--2392, Austin, Texas. Association for Computational Linguistics.

\bibitem[{Rosenberg and Hirschberg(2007)}]{rosenberg-hirschberg-2007-v}
Andrew Rosenberg and Julia Hirschberg. 2007.
\newblock \href {https://aclanthology.org/D07-1043/} {{V}-measure: A conditional entropy-based external cluster evaluation measure}.
\newblock In \emph{Proceedings of the 2007 Joint Conference on Empirical Methods in Natural Language Processing and Computational Natural Language Learning ({EMNLP}-{C}o{NLL})}, pages 410--420, Prague, Czech Republic. Association for Computational Linguistics.

\bibitem[{Shazeer(2020)}]{shazeer-etal-2020-glu}
Noam Shazeer. 2020.
\newblock \href {https://doi.org/10.48550/arXiv.2002.05202} {{{GLU Variants Improve Transformer}}}.
\newblock \emph{Preprint}, arXiv:2002.05202.

\bibitem[{Song and Zhong(2024)}]{song-etal-2024-uncovering}
Jiajun Song and Yiqiao Zhong. 2024.
\newblock \href {https://doi.org/10.48550/arXiv.2310.04861} {Uncovering hidden geometry in {{Transformers}} via disentangling position and context}.
\newblock \emph{Preprint}, arXiv:2310.04861.

\bibitem[{Su et~al.(2023)Su, Lu, Pan, Murtadha, Wen, and Liu}]{su-etal-2023-roformer}
Jianlin Su, Yu~Lu, Shengfeng Pan, Ahmed Murtadha, Bo~Wen, and Yunfeng Liu. 2023.
\newblock \href {https://doi.org/10.48550/arXiv.2104.09864} {{{RoFormer}}: {{Enhanced Transformer}} with {{Rotary Position Embedding}}}.
\newblock \emph{Preprint}, arXiv:2104.09864.

\bibitem[{Urrutia et~al.(2025)Urrutia, Salas, Kozachinskiy, Calderon, Pasten, and Rojas}]{urrutia2025decouplingpositionalsymbolicattention}
Felipe Urrutia, Jorge Salas, Alexander Kozachinskiy, Cristian~Buc Calderon, Hector Pasten, and Cristobal Rojas. 2025.
\newblock \href {https://arxiv.org/abs/2511.11579} {Decoupling positional and symbolic attention behavior in transformers}.
\newblock \emph{Preprint}, arXiv:2511.11579.

\bibitem[{Vaswani et~al.(2017)Vaswani, Shazeer, Parmar, Uszkoreit, Jones, Gomez, Kaiser, and Polosukhin}]{vaswani-etal-2017-attention}
Ashish Vaswani, Noam Shazeer, Niki Parmar, Jakob Uszkoreit, Llion Jones, Aidan~N Gomez, {\L}ukasz Kaiser, and Illia Polosukhin. 2017.
\newblock Attention is all you need.
\newblock In \emph{Advances in Neural Information Processing Systems}, volume~30. Curran Associates, Inc.

\bibitem[{Waldis et~al.(2024)Waldis, Perlitz, Choshen, Hou, and Gurevych}]{waldisetal2024holmes}
Andreas Waldis, Yotam Perlitz, Leshem Choshen, Yufang Hou, and Iryna Gurevych. 2024.
\newblock \href {https://doi.org/10.1162/tacl_a_00718} {Holmes: {{A Benchmark}} to {{Assess}} the {{Linguistic Competence}} of {{Language Models}}}.
\newblock \emph{Transactions of the Association for Computational Linguistics}, 12:1616--1647.

\bibitem[{Wang et~al.(2018)Wang, Singh, Michael, Hill, Levy, and Bowman}]{wang-etal-2018-glue}
Alex Wang, Amanpreet Singh, Julian Michael, Felix Hill, Omer Levy, and Samuel~R. Bowman. 2018.
\newblock \href {https://doi.org/10.18653/v1/W18-5446} {{GLUE}: A multi-task benchmark and analysis platform for natural language understanding}.
\newblock In \emph{Proceedings of the 2018 {EMNLP} Workshop {B}lackbox{NLP}: Analyzing and Interpreting Neural Networks for {NLP}}, pages 353--355, Brussels, Belgium. Association for Computational Linguistics.

\bibitem[{Wang and Chen(2020)}]{wang-chen-2020-position}
Yu-An Wang and Yun-Nung Chen. 2020.
\newblock \href {https://doi.org/10.18653/v1/2020.emnlp-main.555} {What do position embeddings learn? an empirical study of pre-trained language model positional encoding}.
\newblock In \emph{Proceedings of the 2020 Conference on Empirical Methods in Natural Language Processing (EMNLP)}, pages 6840--6849, Online. Association for Computational Linguistics.

\bibitem[{Wu et~al.(2024)Wu, Wang, Xiao, Peng, and Fu}]{wu-etal-2024-retrieval}
Wenhao Wu, Yizhong Wang, Guangxuan Xiao, Hao Peng, and Yao Fu. 2024.
\newblock \href {https://doi.org/10.48550/arXiv.2404.15574} {Retrieval {{Head Mechanistically Explains Long-Context Factuality}}}.
\newblock \emph{Preprint}, arXiv:2404.15574.

\bibitem[{Wu et~al.(2025)Wu, Deshmukh, Wu, Lin, and Mou}]{wu-etal-2025-emergence}
Zijun Wu, Anup~Anand Deshmukh, Yongkang Wu, Jimmy Lin, and Lili Mou. 2025.
\newblock \href {https://doi.org/10.1162/coli_a_00545} {The emergence of chunking structures with hierarchical {RNN}}.
\newblock \emph{Computational Linguistics}, 51(3):815--841.

\bibitem[{Zhang and Sennrich(2019)}]{Zhang-etal-2019-root}
Biao Zhang and Rico Sennrich. 2019.
\newblock Root mean square layer normalization.
\newblock In \emph{Advances in Neural Information Processing Systems}, volume~32. Curran Associates, Inc.

\bibitem[{Zheng et~al.(2024)Zheng, Gao, Shi, Huang, Li, Xiong, Ren, Ng, Jiang, Li et~al.}]{zheng-etal-2024-dape}
Chuanyang Zheng, Yihang Gao, Han Shi, Minbin Huang, Jingyao Li, Jing Xiong, Xiaozhe Ren, Michael Ng, Xin Jiang, Zhenguo Li, and 1 others. 2024.
\newblock Dape: Data-adaptive positional encoding for length extrapolation.
\newblock \emph{Advances in Neural Information Processing Systems}, 37:26659--26700.

\bibitem[{Zheng et~al.(2025)Zheng, Gao, Shi, Xiong, Sun, Li, Huang, Ren, Ng, Jiang, Li, and Li}]{zheng-etal-2025-dape}
Chuanyang Zheng, Yihang Gao, Han Shi, Jing Xiong, Jiankai Sun, Jingyao Li, Minbin Huang, Xiaozhe Ren, Michael Ng, Xin Jiang, Zhenguo Li, and Yu~Li. 2025.
\newblock \href {https://doi.org/10.18653/v1/2025.acl-long.522} {{DAPE} v2: Process attention score as feature map for length extrapolation}.
\newblock In \emph{Proceedings of the 63rd Annual Meeting of the Association for Computational Linguistics (Volume 1: Long Papers)}, pages 10628--10666, Vienna, Austria. Association for Computational Linguistics.

\end{thebibliography}

\appendix
\section{DSTG-NeoBERT Architecture Details}
\subsection{T5 Positional Bias}
\label{app:t5-positional-bias}
This section provides details on the T5 bucketed bias used in DSTG-NeoBERT's relative positional component (Section~\ref{sec:posbert_architecture}). T5 \cite{raffel-etal-2020-exploring} introduces a learned relative position bias inside self-attention. For a query token at position $i$ and a key token at position $j$, the model considers their relative distance and associates it with a learned scalar bias, which is learned per attention head. In fact, T5 does not learn one parameter for every possible distance; distances are first mapped to a finite set of buckets. Nearby distances are represented more finely, while larger distances are grouped together more coarsely. T5 uses $32$ relative-position embeddings (i.e.\ $32$ buckets), with bucket ranges that grow logarithmically up to a relative offset of $128$; beyond that, all larger distances are mapped to the same bucket. This means that small offsets such as one or two positions apart can receive distinct learned biases, whereas larger offsets such as $40$, $50$, or $60$ tokens apart may share the same bucket, and any offset larger than $128$ is treated identically from the viewpoint of a single layer.

These learned scalar biases are added directly to the attention logits before the softmax. Using the usual attention notation, this can be written as
\[
\mathrm{Attention}(Q,K,V)
=
\mathrm{softmax}\!\left(\frac{QK^\top}{\sqrt{d_k}} + B\right)V,
\]
where $B \in \mathbb{R}^{n \times n}$ is the matrix of relative position biases, $n$ is the sequence length, and each entry $B_{ij}$ depends on the bucketed relative distance between query position $i$ and key position $j$ for the corresponding attention head.

\subsection{Attention Process}
\label{subapp:attention}
\begin{figure*}[t]
 \centering
  \includegraphics[width=0.9\linewidth]{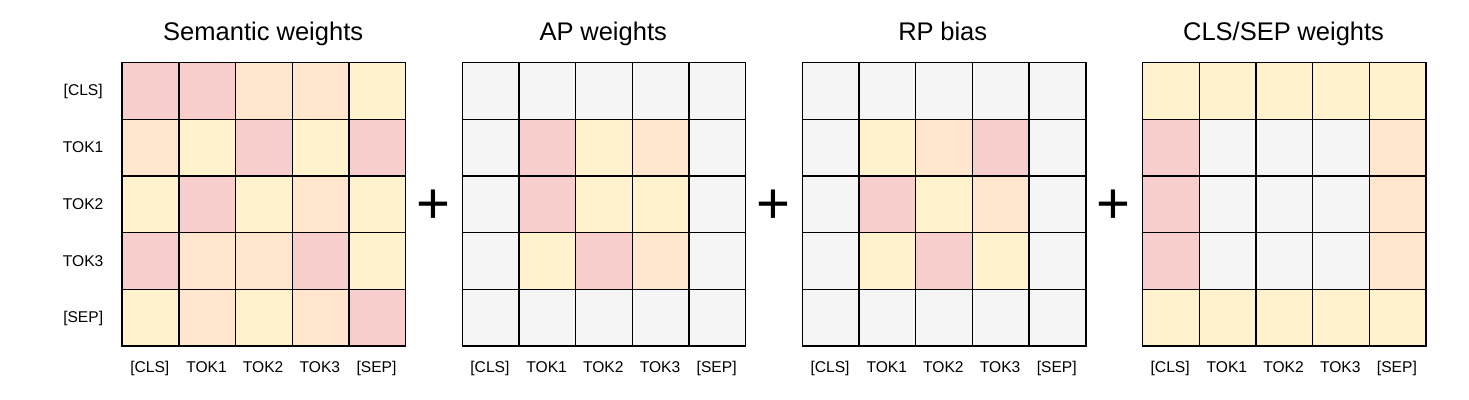}
  \caption{DSTG-NeoBERT attention weights mechanism. Grayed-out squares correspond to discarded weights. From left to right: (1) The semantic space can match all tokens. (2) The AP space can match all but [CLS] and [SEP] tokens. (3) The RP bias learns a per-bucket weight (all buckets are of size 1 in this example), [CLS] and [SEP] are discarded. Notice that the weights are not symmetrical nor necessarily decaying with distance. (4) The learned $\rho$ weights for [CLS] and [SEP] applied to the sum.}
  \label{fig:attn_weights_sum}
\end{figure*}

Figure~\ref{fig:attn_weights_sum} displays a graphical example of the attention logits computation described in Section~\ref{sec:posbert_architecture}.

\section{Complementary Analysis of Learned Representations}


\subsection{Comparison of Learned AP Embeddings}
\label{subapp:learned_ap_embeddings}
\begin{figure}
 \centering
  \includegraphics[width=0.99\linewidth]{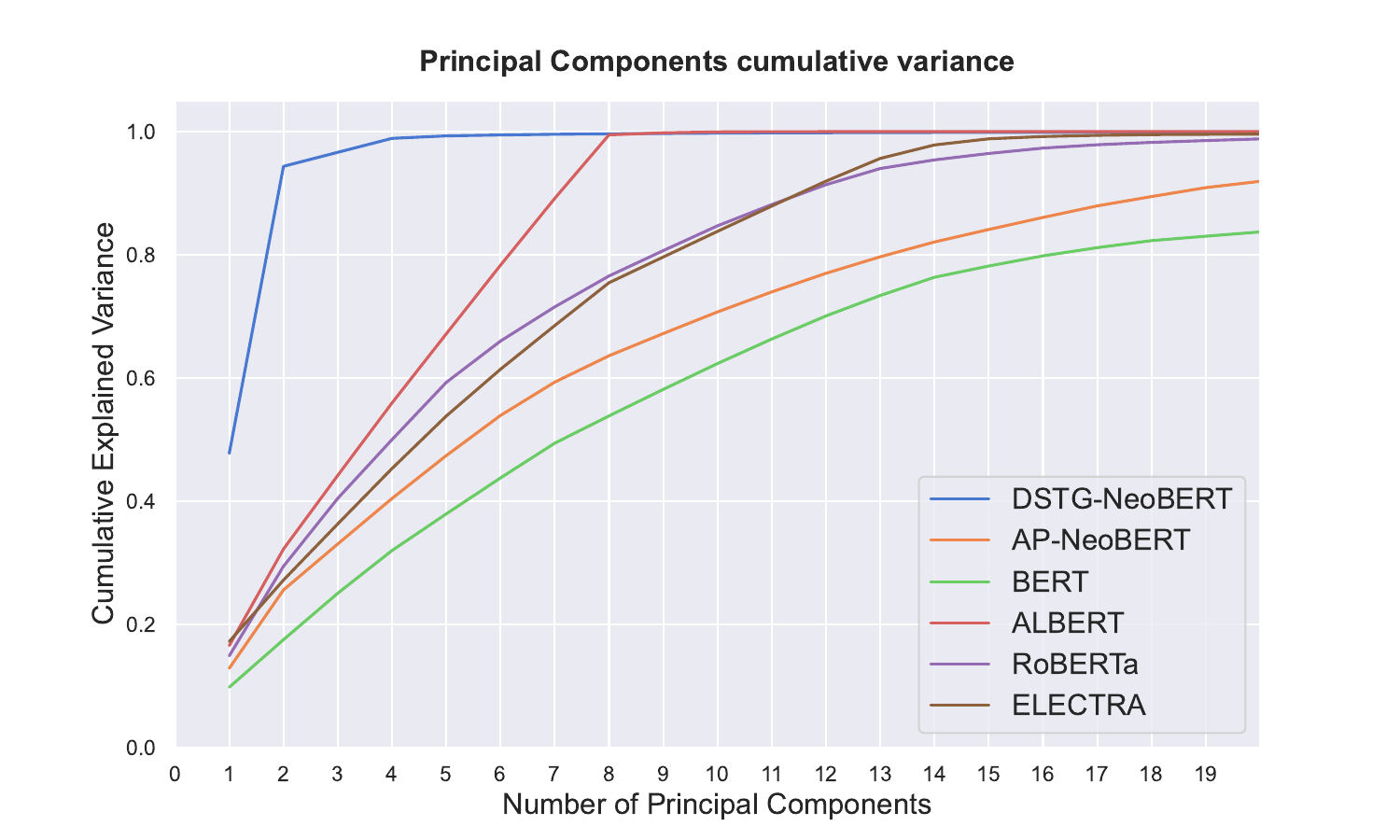}
  \caption{Cumulative explained variance of singular vectors for different encoder-based models with learned AP embeddings.}
  \label{fig:svd_learned_embed}
\end{figure}

\begin{figure*}
\centering
\begin{subfigure}{0.45\textwidth}
    \includegraphics[width=\textwidth]{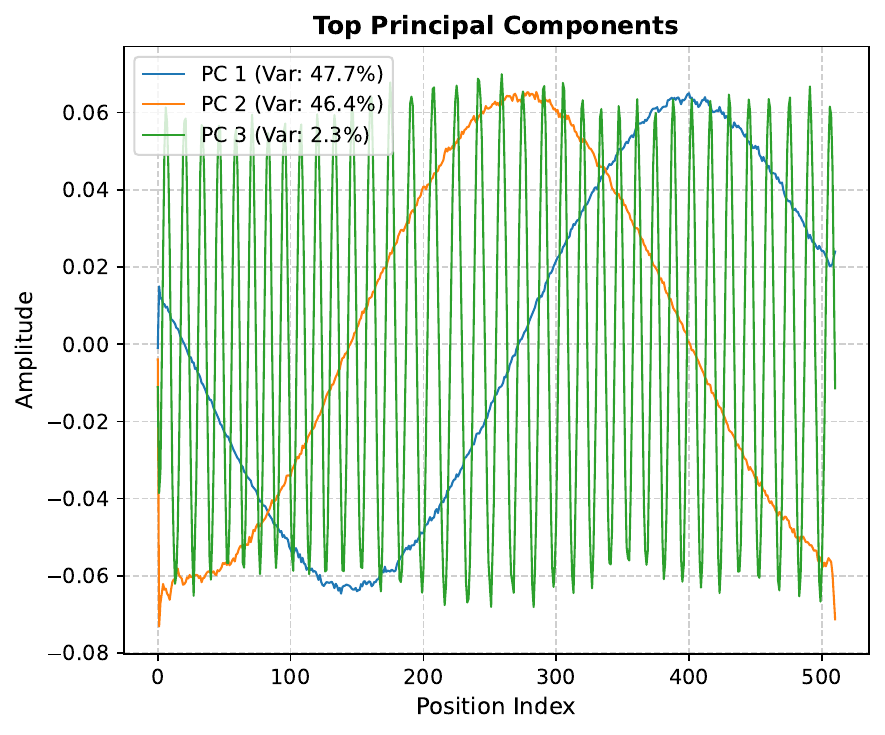}
    \caption{DSTG-NeoBERT}
    \label{fig:DSTG-NeoBERT_acp}
\end{subfigure}
\begin{subfigure}{0.45\textwidth}
    \includegraphics[width=\textwidth]{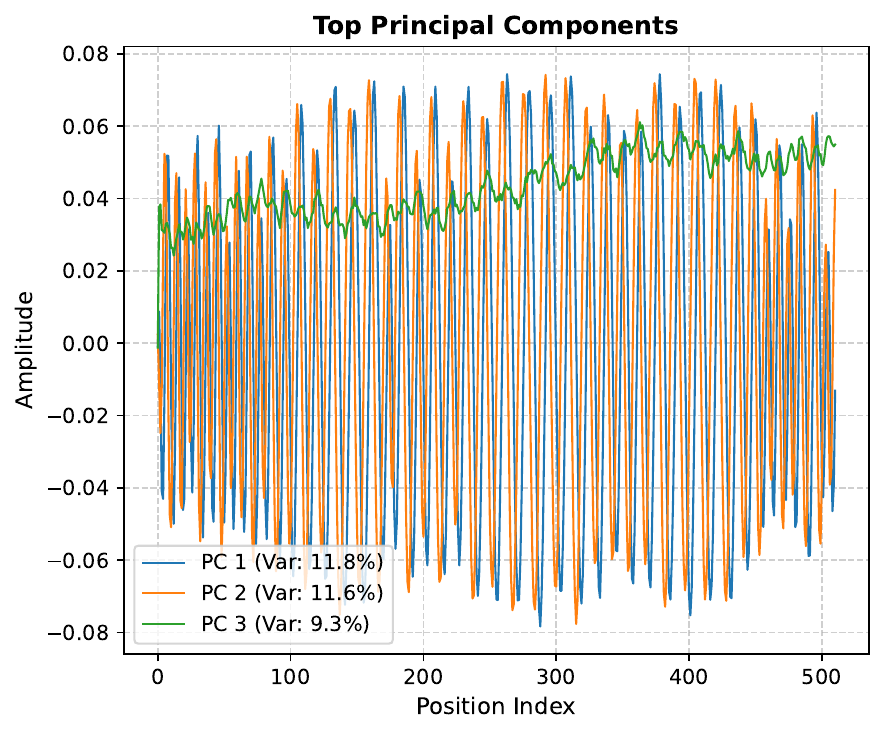}
    \caption{AP-NeoBERT}
    \label{fig:ap-neobert_acp}
\end{subfigure}
\begin{subfigure}{0.45\textwidth}
    \includegraphics[width=\textwidth]{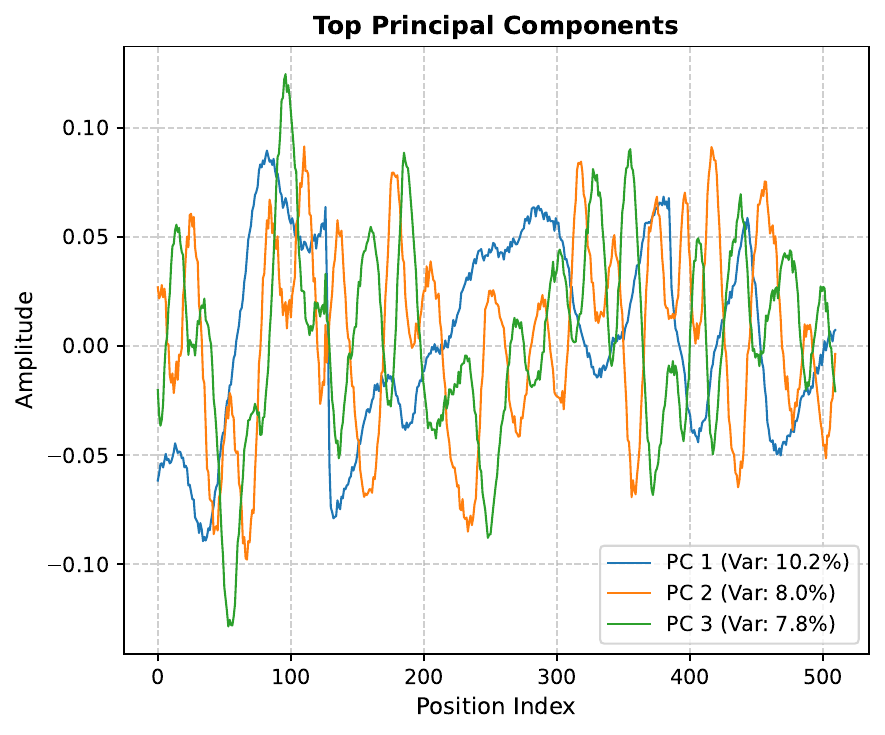}
    \caption{BERT}
    \label{fig:bert_acp}
\end{subfigure}
\begin{subfigure}{0.45\textwidth}
    \includegraphics[width=\textwidth]{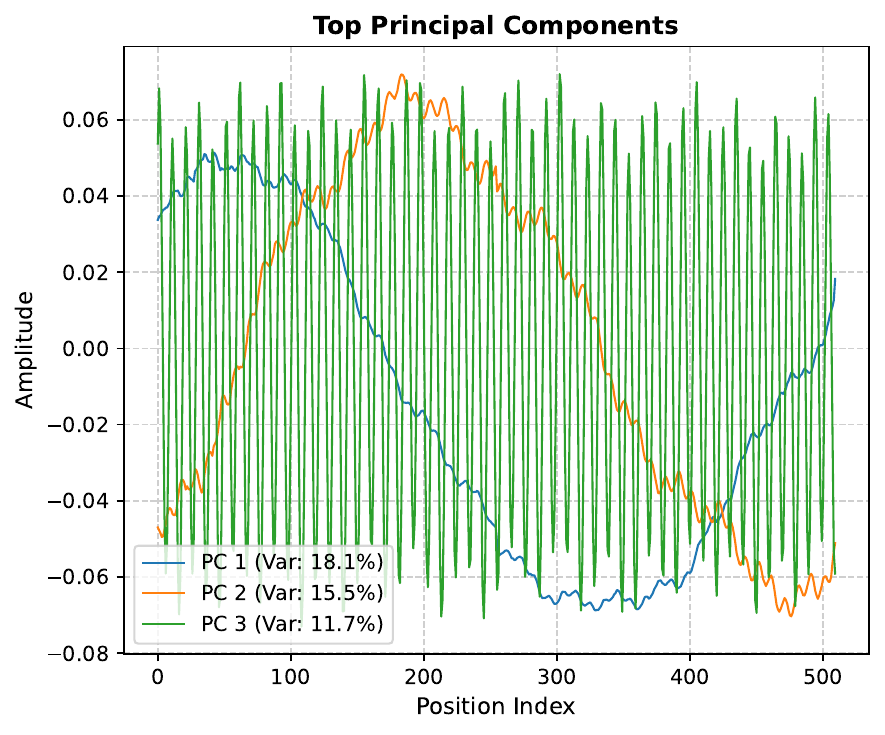}
    \caption{ALBERT}
    \label{fig:albert_acp}
\end{subfigure}
\begin{subfigure}{0.45\textwidth}
    \includegraphics[width=\textwidth]{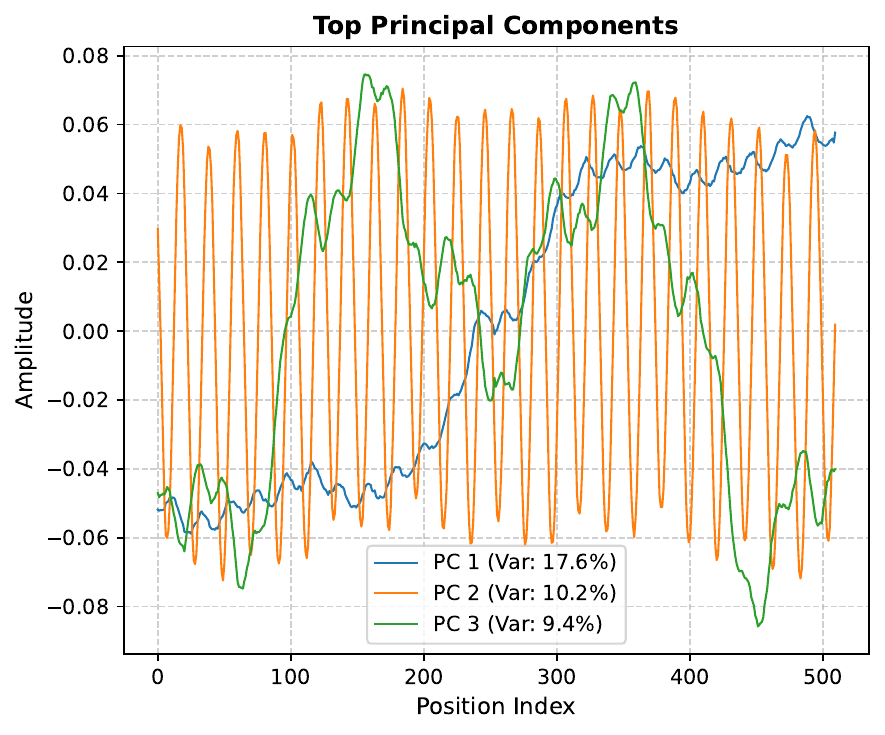}
    \caption{ELECTRA}
    \label{fig:electra_acp}
\end{subfigure}
\begin{subfigure}{0.45\textwidth}
    \includegraphics[width=\textwidth]{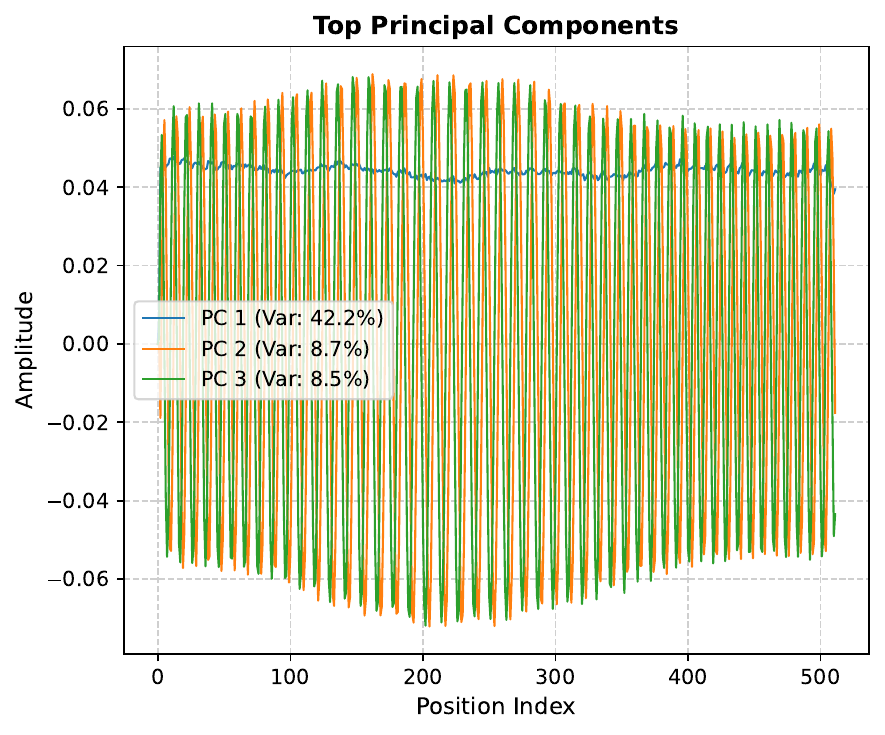}
    \caption{RoBERTa}
    \label{fig:roberta_acp}
\end{subfigure}

\caption{The first three PCs of the AP embeddings of different models.}
\label{fig:models_PCs_acp}
\end{figure*}

This section extends the PCA and DCT analysis of Section~\ref{sec:analysis} by comparing DSTG-NeoBERT's learned AP embeddings to those of AP-NeoBERT and other encoder-based Transformers with learned AP embeddings: BERT \cite{devlin-etal-2019-bert}, ALBERT \cite{lan-etal-2020-albert}, RoBERTa \cite{liu-etal-2019-roberta}, and ELECTRA \cite{clark-etal-2020-electra}. Figure~\ref{fig:svd_learned_embed} shows the cumulative variance of Principal Components (PCs) for each model. DSTG-NeoBERT is significantly lower-dimensional than other systems. The principal components of each model are displayed in Figure~\ref{fig:models_PCs_acp}. DSTG-NeoBERT displays strongly sinusoidal patterns with a much higher explained variance.

\subsection{Additional Probing Experiments}
\label{subapp:intermodel}

\begin{figure*}
\centering
\begin{subfigure}{0.49\textwidth}
    \includegraphics[width=\textwidth]{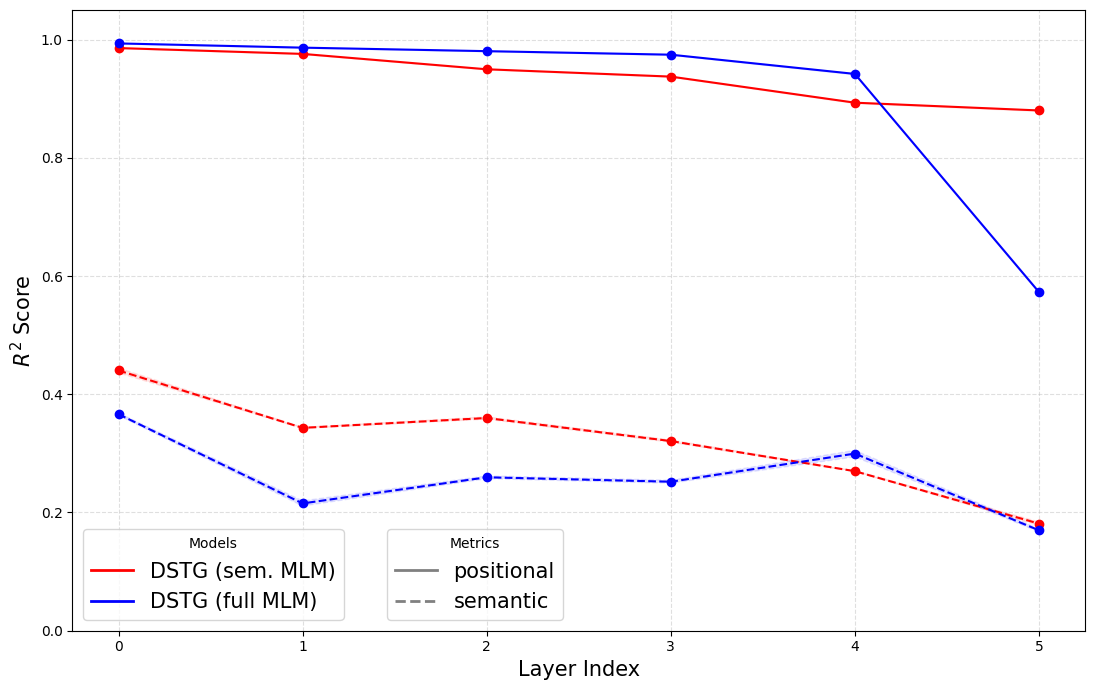}
    \caption{token-level AP probe}
\end{subfigure}
\begin{subfigure}{0.49\textwidth}
    \includegraphics[width=\textwidth]{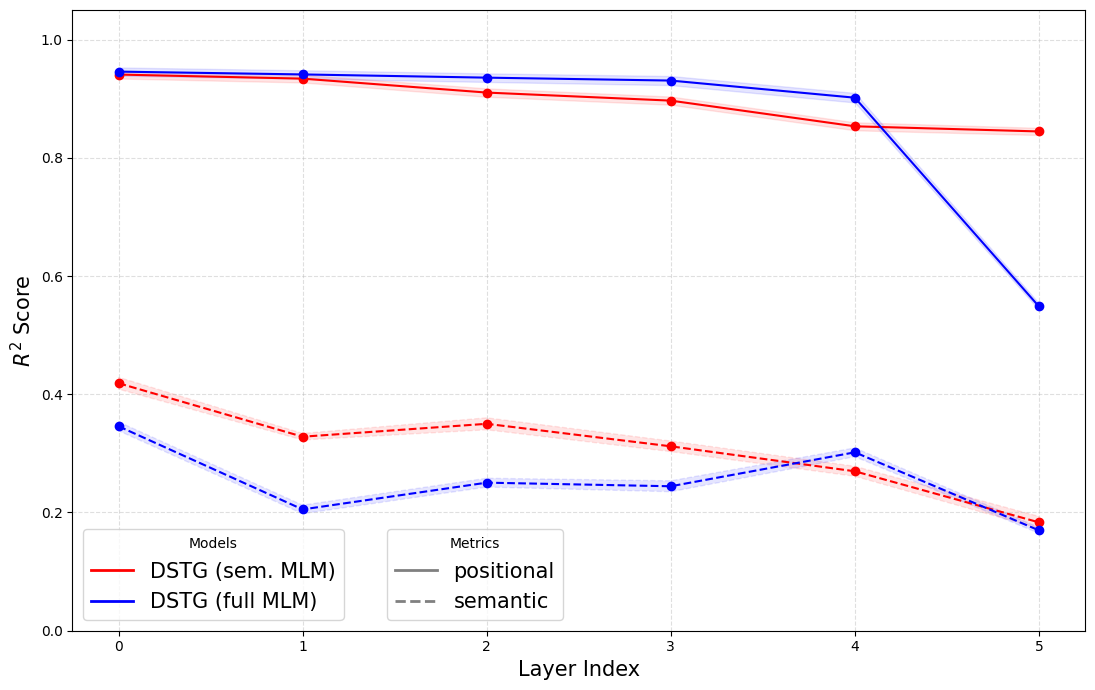}
    \caption{segment-level AP probe}
\end{subfigure}
\caption{Results of the structural probes on DSTG-NeoBERT with MLM on semantic only (red) and DSTG-NeoBERT with MLM on both semantic and positional (blue), for their positional (solid line) and semantic (dotted line) embeddings.}
\label{fig:DSTG_probes}
\end{figure*}

\paragraph{MLM on the Full $d_{model}$ Embedding}
To further confirm that the MLM training objective is the cause of the loss in structural information, we trained a disentangled architecture with the MLM objective applied to both the positional and the semantic spaces, and trained the structural probes. We compare AP-NeoBERT, DSTG-NeoBERT (semantic MLM) and DSTG-NeoBERT (full embedding MLM), and provide the results for the token-level AP and segment-level AP in Figure~\ref{fig:DSTG_probes}. Note that we also provide the probing results on the last positional layer of DSTG-NeoBERT (semantic MLM), which is never used during training. We observe a significant loss of positional information in the last layer when MLM is applied to the full embedding, confirming that the loss of positional information stems from the MLM objective.

\begin{figure*}
\centering
\begin{subfigure}{0.49\textwidth}
    \includegraphics[width=\textwidth]{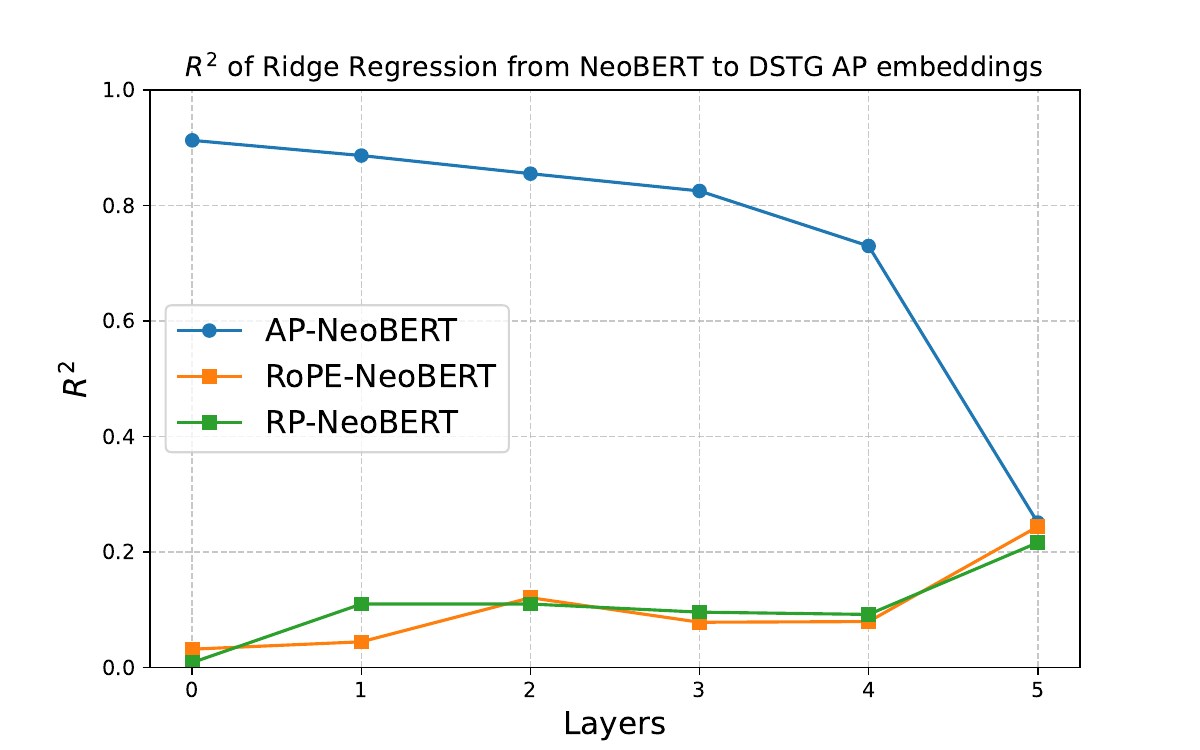}
    \caption{regression to positional embedding}
\end{subfigure}
\begin{subfigure}{0.49\textwidth}
    \includegraphics[width=\textwidth]{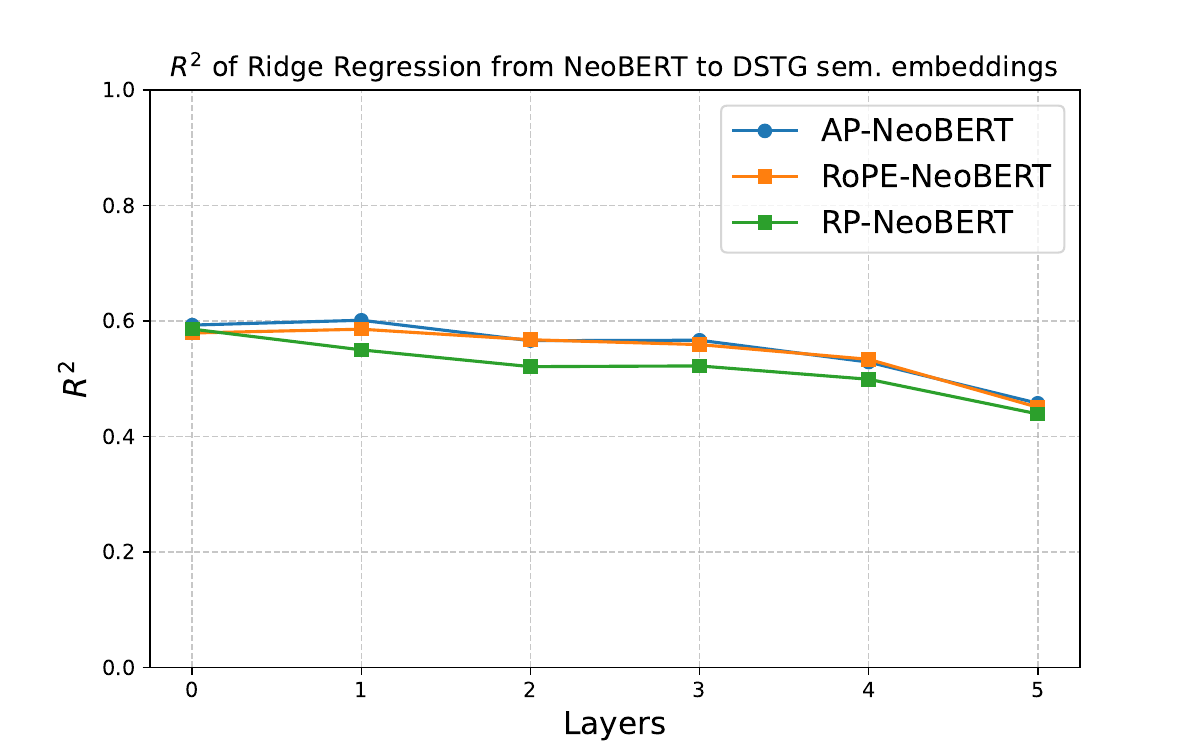}
    \caption{regression to semantic embedding}
\end{subfigure}
\caption{Regression from NeoBERT variants to DSTG-NeoBERT subspaces.}
\label{fig:reg_to_posbert}
\end{figure*}

\paragraph{Inter-Model Probing}
To evaluate how much of the information encoded by DSTG-NeoBERT is represented in the three other baselines, we regress the hidden states of each baseline (AP, RP and RoPE) onto the disentangled AP and semantic representations of DSTG-NeoBERT. Results are shown in Figure~\ref{fig:reg_to_posbert}. While only AP-NeoBERT encodes the positional information, all three models achieve comparable regression $R^2$ when predicting the semantic representations of DSTG-NeoBERT. Given that the probe is linear, it is not expected to achieve an $R^2$ score close to 1 on the semantic embeddings probe.

\begin{table}
    \centering
    \small
    \begin{tabular}{lcccc|c}
        \toprule
        \textbf{Model} & \textbf{AP} & \textbf{RP} & \multicolumn{2}{c|}{\textbf{RoPE}} & \textbf{DSTG} \\
        \cmidrule(lr){2-2} \cmidrule(lr){3-3} \cmidrule(lr){4-5} \cmidrule(lr){6-6}
        Hidden Size & 720 & 720 &  720 &  768 & $720+48$\\
        \midrule
        sentence ord. & 0.72 & 0.72 & 0.65 & 0.63 & \textbf{0.89} \\
        top-constit. & 0.35 & 0.35 & 0.37 & 0.42 & \textbf{0.47} \\
        POS tagging & 0.62 & 0.62 & 0.63 & 0.60 & \textbf{0.71} \\
        Island effect & 0.73 & 0.73 & 0.72 & 0.77 & \textbf{0.79} \\
        Readability & 0.22 & 0.22 & 0.20 & 0.18 & \textbf{0.27} \\
        \bottomrule
    \end{tabular}
    \caption{Linguistic Phenomena of Flash-Holmes with the best improvements (over RoPE-720). The first four rows are F1 score. The last row is Pearson correlation.}
    \label{tab:holmes_results_best}
\end{table}

\section{Experiment Details}
\label{app:experiments}
\subsection{GLUE}
\label{subapp:glue}
\paragraph{Metrics} 
Following \citet{wang-etal-2018-glue}, we use different metrics for the different datasets. For STSB, we provide Pearson correlation. For CoLA, Matthews correlation. For QQP, the F1 score. The accuracy is given for all other tasks.

\paragraph{Hyperparameters and Training} We run the hyperparameter search with learning rates in $\{6e\text{-}6, 5e\text{-}6, 2e\text{-}5\}$, batch sizes in $\{4, 16, 32\}$, and weight decays in $\{1e\text{-}2, 1e\text{-}5\}$. Following NeoBERT, we finetune on the training split and evaluate on the validation split every $n = \text{min}(500, \text{len(dataloader)} // 10)$ steps. Training is stopped after 15 evaluations without improvement, or after 10 epochs. Following standard practice, WNLI is excluded from the evaluation. For each model, we report only the performance of the best hyperparameter combination. Results are displayed in Table~\ref{tab:results_glue}.

\begin{table*}
    \centering
    \small
    \begin{tabular}{lcccccccc|c}
        \toprule
        \textbf{Model} & \textbf{MNLI} & \textbf{QNLI} & \textbf{QQP} & \textbf{RTE} & \textbf{SST} & \textbf{MRPC} & \textbf{coLA} & \textbf{STS} & \textbf{Avg.} \\
        \midrule
        AP-NeoBERT & 0.80 & 0.88 & 0.86 & \textbf{0.68} & 0.92 & \textbf{0.85} & 0.45 & \textbf{0.87} & \textbf{0.79}\\
        RP-NeoBERT & 0.79 & 0.87 & 0.86 & 0.61 & 0.90 & 0.82 & 0.36 & 0.85 & 0.75 \\
        RoPE-NeoBERT & \textbf{0.81} & \textbf{0.89} & \textbf{0.87} & 0.67 & \textbf{0.93} & \textbf{0.85} & \textbf{0.48} & \textbf{0.87} & \textbf{0.79}\\
        DSTG-NeoBERT & \textbf{0.81} & 0.87 & \textbf{0.87} & 0.60 & 0.92 & \textbf{0.85} & \textbf{0.48} & \textbf{0.87} & 0.78 \\
        \bottomrule
    \end{tabular}
    \caption{Results on the GLUE Benchmark}
    \label{tab:results_glue}
\end{table*}

\subsection{MTEB}
\label{subapp:mteb}
\paragraph{Metrics} Similar to GLUE, metrics differ for each task. We use the metrics recommended in \citet{muennighoff-etal-2023-mteb}. For classification, the main metric is accuracy. For clustering, the main metric is the v-measure \cite{rosenberg-hirschberg-2007-v}. For pair classification, it is the average precision. For reranking, MAP is used. Retrieval uses nDCG@10. STS and summarization use Spearman correlation. Results are displayed in Table~\ref{tab:results_mtev}.

\begin{table*}
    \centering
    \small
    \begin{tabular}{lccccccc|c}
        \toprule
        \textbf{Model} & \textbf{Class.} & \textbf{Clust.} & \textbf{PairClass.} & \textbf{Rerank.} & \textbf{Retri.} & \textbf{STS} & \textbf{Summ} & \textbf{Avg.} \\
        \midrule
        AP-NeoBERT & \textbf{0.63} &\textbf{ 0.31} & \textbf{0.73} & \textbf{0.52} & \textbf{0.15} & \textbf{0.67} & 0.28 & \textbf{0.47} \\
        RP-NeoBERT & 0.59 & 0.30 & 0.65 & 0.50 & 0.12 & 0.61 & 0.29 & 0.43  \\
        RoPE-NeoBERT & \textbf{0.63} & \textbf{0.31} & 0.70 & 0.51 & 0.13 & 0.65 & \textbf{0.30} & 0.46 \\
        DSTG-NeoBERT & \textbf{0.63} & 0.30 & 0.69 & \textbf{0.52} & 0.14 & 0.63 & \textbf{0.30} & 0.46  \\
        \bottomrule
    \end{tabular}
    \caption{Results on the MTEB Benchmark}
    \label{tab:results_mtev}
\end{table*}

\subsection{Flash-Holmes}
\label{subapp:holmes}
\paragraph{Results} A deeper analysis of the results on Flash-Holmes shows that the highest gains are found in phenomena related to macroscopic structure. Table~\ref{tab:holmes_results_best} shows the results for the five phenomena with the highest gains. \textit{top-constituent} prediction probes highest-level grammatical phrases (subject, verb...). \textit{Island effect} refers to why it is ungrammatical to extract or move a word or phrase out of certain complex structural domains, known as ``islands,'' to form questions or relative clauses. \textit{Readability} computes the Flesch Score \citep{flesch1948new}, a measure of readability based on the number of words in the sentence and the average number of syllables per word.
We also provide the results on all phenomena, per subfield: Discourse (Table~\ref{tab:holmes_results_discourse}), Morphology (Table~\ref{tab:holmes_results_morphology}), Reasoning (Table~\ref{tab:holmes_results_reasoning}), Semantics (Table~\ref{tab:holmes_results_semantics}) and Syntax (Table~\ref{tab:holmes_results_syntax}). We find a clear advantage of the disentangled architecture on the Syntax, Semantics and Discourse subfields. Refer to \citet{waldisetal2024holmes} for the definition of each phenomenon.

We discarded 7 out of the 215 probing datasets used in Flash-Holmes because the results were statistically insignificant (standard deviation $> 0.1$) for at least one model: blimp-determiner\_noun\_agreement\_irregular\_2, protoroles-changes\_possession, protoroles-exists\_as\_physical, protoroles-location\_of\_event, protoroles-makes\_physical\_contact, protoroles-predicate\_changed\_argument and protoroles-stationary. Remaining datasets have a mean standard deviation of 0.02.



\begin{table*}[htbp]
    \centering
    \begin{filecontents*}{"tables/holmes_results_discourse.csv"}
subfield,phenomena,AP,DSTG,RP,RoPE-720,RoPE-768
discourse,bridging,0.49,\textbf{0.50},0.49,0.49,0.49
discourse,coreference resolution,0.66,\textbf{0.69},0.66,0.67,0.68
discourse,discourse connective,0.08,\textbf{0.10},0.08,0.07,0.09
discourse,next sentence prediction,0.44,0.43,0.44,\textbf{0.45},0.44
discourse,rethorical structure theory,0.19,\textbf{0.22},0.19,0.19,0.19
discourse,sentence order,0.72,\textbf{0.89},0.72,0.65,0.63
\end{filecontents*}
\csvreader[
        tabular=l c c c c | c, 
        table head=\toprule 
             \textbf{Phenomena} & 
            \textbf{AP} & \textbf{RP} & \textbf{RoPE-720} & \textbf{RoPE-768} & \textbf{DSTG} \\ \midrule,
        late after line=\\,
        late after last line=\\\bottomrule
    ]{"tables/holmes_results_discourse.csv"}
    {
        phenomena=\phenomena,
        AP=\AP,
        RP=\RP,
        RoPE-720=\ropeA, 
        RoPE-768=\ropeB,
        DSTG=\DSTG
    }
    {
         \phenomena & \AP  & \RP & \ropeA & \ropeB & \DSTG
    }
    \caption{Flash-Holmes Results on the \textbf{Discourse} subfield}
    \label{tab:holmes_results_discourse}
\end{table*}

\begin{table*}[htbp]
    \centering
    \begin{filecontents*}{"tables/holmes_results_morphology.csv"}
subfield,phenomena,AP,DSTG,RP,RoPE-720,RoPE-768
morphology,anaphor agreement,0.68,0.65,0.68,0.65,\textbf{0.69}
morphology,determiner noun agreement,0.54,0.54,0.54,0.55,\textbf{0.60}
morphology,irregular forms,0.80,\textbf{0.86},0.80,0.80,0.81
morphology,subject-verb agreement,0.54,0.58,0.54,0.54,\textbf{0.59}
\end{filecontents*}
\csvreader[
        tabular=l c c c c | c, 
        table head=\toprule 
             \textbf{Phenomena} & 
            \textbf{AP} & \textbf{RP} & \textbf{RoPE-720} & \textbf{RoPE-768} & \textbf{DSTG} \\ \midrule,
        late after line=\\,
        late after last line=\\\bottomrule
    ]{"tables/holmes_results_morphology.csv"}
    {
        phenomena=\phenomena,
        AP=\AP,
        RP=\RP,
        RoPE-720=\ropeA, 
        RoPE-768=\ropeB,
        DSTG=\DSTG
    }
    {
         \phenomena & \AP  & \RP & \ropeA & \ropeB & \DSTG
    }
    \caption{Flash-Holmes Results on the \textbf{Morphology} subfield}
    \label{tab:holmes_results_morphology}
\end{table*}

\begin{table*}[htbp]
    \centering
    \begin{filecontents*}{"tables/holmes_results_reasoning.csv"}
subfield,phenomena,AP,DSTG,RP,RoPE-720,RoPE-768
reasoning,age comparison ,0.42,0.41,0.42,\textbf{0.46},0.43
reasoning,always never,\textbf{0.44},\textbf{0.44},\textbf{0.44},\textbf{0.44},\textbf{0.44}
reasoning,antonym negation,\textbf{0.52},0.51,\textbf{0.52},0.50,0.51
reasoning,encyclopedic composition,0.40,\textbf{0.41},0.40,0.40,\textbf{0.41}
reasoning,multi-hop composition,\textbf{0.40},\textbf{0.40},\textbf{0.40},\textbf{0.40},\textbf{0.40}
reasoning,negation,0.52,\textbf{0.53},0.52,0.51,0.51
reasoning,object comparision,0.48,\textbf{0.49},0.48,0.47,0.45
reasoning,property conjunction,0.40,0.40,0.40,\textbf{0.41},\textbf{0.41}
reasoning,speculation,0.44,\textbf{0.46},0.44,0.40,0.44
reasoning,taxonomy conjunction,\textbf{0.41},0.40,\textbf{0.41},0.40,0.40
\end{filecontents*}
\csvreader[
        tabular=l c c c c | c, 
        table head=\toprule 
             \textbf{Phenomena} & 
            \textbf{AP} & \textbf{RP} & \textbf{RoPE-720} & \textbf{RoPE-768} & \textbf{DSTG} \\ \midrule,
        late after line=\\,
        late after last line=\\\bottomrule
    ]{"tables/holmes_results_reasoning.csv"}
    {
        phenomena=\phenomena,
        AP=\AP,
        RP=\RP,
        RoPE-720=\ropeA, 
        RoPE-768=\ropeB,
        DSTG=\DSTG
    }
    {
         \phenomena & \AP  & \RP & \ropeA & \ropeB & \DSTG
    }
    \caption{Flash-Holmes Results on the \textbf{Reasoning} subfield}
    \label{tab:holmes_results_reasoning}
\end{table*}

\begin{table*}[htbp]
    \centering
    \begin{filecontents*}{"tables/holmes_results_semantics.csv"}
subfield,phenomena,AP,DSTG,RP,RoPE-720,RoPE-768
semantics,complex words,0.58,\textbf{0.62},0.58,0.59,0.55
semantics,coordination inversion,0.52,\textbf{0.57},0.52,0.55,0.55
semantics,event structure,\textbf{0.45},0.44,\textbf{0.45},\textbf{0.45},0.44
semantics,factuality,\textbf{0.46},0.45,\textbf{0.46},0.45,\textbf{0.46}
semantics,genericity,0.13,\textbf{0.16},0.13,0.14,0.13
semantics,metaphor,0.68,\textbf{0.69},0.68,\textbf{0.69},0.68
semantics,named entity labeling,0.53,\textbf{0.59},0.53,0.55,0.54
semantics,negative polarity item licensing,0.91,\textbf{0.98},0.91,0.93,0.95
semantics,object animacy,0.79,0.85,0.79,0.81,\textbf{0.87}
semantics,object gender,0.60,0.55,0.60,0.62,\textbf{0.67}
semantics,passive,0.57,\textbf{0.59},0.57,0.55,\textbf{0.59}
semantics,protoroles,0.11,\textbf{0.16},0.11,0.14,0.10
semantics,quantifiers,0.91,\textbf{0.95},0.91,0.91,0.90
semantics,relation classification,0.34,\textbf{0.39},0.34,0.37,0.37
semantics,semantic odd man out,0.54,\textbf{0.55},0.54,\textbf{0.55},0.54
semantics,sentiment analysis ,0.24,\textbf{0.30},0.24,0.27,0.24
semantics,subject animacy,0.86,0.85,0.86,\textbf{0.90},\textbf{0.90}
semantics,subject gender,0.72,0.57,0.72,\textbf{0.75},0.68
semantics,synonym-/antonym-detection,0.63,\textbf{0.64},0.63,0.63,0.63
semantics,tense,\textbf{0.91},0.88,\textbf{0.91},\textbf{0.91},0.84
semantics,time,\textbf{0.12},0.11,\textbf{0.12},0.11,\textbf{0.12}
semantics,verb dynamic,0.69,\textbf{0.70},0.69,0.67,0.61
semantics,word content,0.09,\textbf{0.13},0.09,0.12,\textbf{0.13}
semantics,word sense,\textbf{0.18},\textbf{0.18},\textbf{0.18},0.15,0.16
\end{filecontents*}
\csvreader[
        tabular=l c c c c | c, 
        table head=\toprule 
             \textbf{Phenomena} & 
            \textbf{AP} & \textbf{RP} & \textbf{RoPE-720} & \textbf{RoPE-768} & \textbf{DSTG} \\ \midrule,
        late after line=\\,
        late after last line=\\\bottomrule
    ]{"tables/holmes_results_semantics.csv"}
    {
        phenomena=\phenomena,
        AP=\AP,
        RP=\RP,
        RoPE-720=\ropeA, 
        RoPE-768=\ropeB,
        DSTG=\DSTG
    }
    {
         \phenomena & \AP  & \RP & \ropeA & \ropeB & \DSTG
    }
    \caption{Flash-Holmes Results on the \textbf{Semantics} subfield}
    \label{tab:holmes_results_semantics}
\end{table*}

\begin{table*}[htbp]
    \centering
    \begin{filecontents*}{"tables/holmes_results_syntax.csv"}
subfield,phenomena,AP,DSTG,RP,RoPE-720,RoPE-768
syntax,anaphor agreement,0.65,0.63,0.65,\textbf{0.66},0.57
syntax,argument structure,0.72,0.73,0.72,0.72,\textbf{0.74}
syntax,bigram-shift,0.76,\textbf{0.78},0.76,0.76,\textbf{0.78}
syntax,binding,0.71,\textbf{0.77},0.71,0.72,0.74
syntax,case,\textbf{0.89},0.72,\textbf{0.89},0.74,0.73
syntax,constituent parsing,0.80,\textbf{0.87},0.80,0.83,0.81
syntax,control / raising,0.81,0.82,0.81,0.80,\textbf{0.83}
syntax,deoncausative-inchoative alternation ,0.77,\textbf{0.81},0.77,0.74,0.70
syntax,ellipsis,0.63,\textbf{0.65},0.63,0.64,0.64
syntax,filler gap,0.85,\textbf{0.94},0.85,0.87,0.90
syntax,island effects,0.73,\textbf{0.79},0.73,0.72,0.77
syntax,local attractor,0.66,0.65,0.66,0.63,\textbf{0.71}
syntax,negative polarity item licensing,0.80,0.82,0.80,0.81,\textbf{0.85}
syntax,object number,0.78,\textbf{0.79},0.78,0.77,\textbf{0.79}
syntax,part-of-speech,0.62,\textbf{0.71},0.62,0.63,0.60
syntax,readability,0.22,\textbf{0.27},0.22,0.20,0.18
syntax,sentence length,\textbf{0.05},\textbf{0.05},\textbf{0.05},\textbf{0.05},\textbf{0.05}
syntax,subject number,0.78,0.79,0.78,0.78,\textbf{0.81}
syntax,subject-verb agreement,0.52,\textbf{0.55},0.52,0.52,\textbf{0.55}
syntax,top-constituent-task,0.35,\textbf{0.47},0.35,0.37,0.42
syntax,tree-depth ,0.22,\textbf{0.27},0.22,0.23,0.22
\end{filecontents*}
\csvreader[
        tabular=l c c c c | c, 
        table head=\toprule 
             \textbf{Phenomena} & 
            \textbf{AP} & \textbf{RP} & \textbf{RoPE-720} & \textbf{RoPE-768} & \textbf{DSTG} \\ \midrule,
        late after line=\\,
        late after last line=\\\bottomrule
    ]{"tables/holmes_results_syntax.csv"}
    {
        phenomena=\phenomena,
        AP=\AP,
        RP=\RP,
        RoPE-720=\ropeA, 
        RoPE-768=\ropeB,
        DSTG=\DSTG
    }
    {
         \phenomena & \AP  & \RP & \ropeA & \ropeB & \DSTG
    }
    \caption{Flash-Holmes Results on the \textbf{Syntax} subfield}
    \label{tab:holmes_results_syntax}
\end{table*}

\subsection{SQuAD}
\label{subapp:squad}
For SQuAD, we report the best results of exact match (EM) and F1-score with a grid search of learning rates in $\{1e\text{-}5, 3e\text{-}5, 5e\text{-}5\}$. For all four models, the best result is found with a learning rate of $3e\text{-}5$.

\end{document}